\documentclass{article}

\usepackage{amssymb,amsfonts,amsmath, bm, mathtools}
\usepackage{dsfont}
\usepackage{tikz}
\usepackage{fullpage}
\usepackage{multirow}
\usepackage{float}  
\usepackage[normalem]{ulem}
\usepackage[
  backend=biber,
  style=numeric,
  sorting=none
]{biblatex}

\usepackage{enumitem}
\usepackage{booktabs}
\usepackage[colorlinks=true, linkcolor=blue, filecolor=blue, citecolor=blue, urlcolor=blue]{hyperref}
\usepackage{cleveref}
\usepackage{graphicx}
\usepackage{xcolor}
\usepackage{tikz}
\usetikzlibrary{shapes.arrows, arrows, decorations.markings, decorations.text}
\usetikzlibrary{calc}
\usepackage{subcaption}
\usepackage{xcolor}
\usepackage{pifont}

\usepackage{authblk}

\newcommand{\cmark}{\textcolor{green!60!black}{\ding{51}}}
\newcommand{\xmark}{\textcolor{red!70!black}{\ding{55}}}

\definecolor{python1}{HTML}{1f77b4}
\definecolor{python2}{HTML}{ff7f0e}
\definecolor{python3}{HTML}{2ca02c}
\definecolor{python4}{HTML}{d62728}
\definecolor{python5}{HTML}{9467bd}
\definecolor{python6}{HTML}{8c564b}
\definecolor{python7}{HTML}{e377c2}
\definecolor{python8}{HTML}{7f7f7f}
\definecolor{python9}{HTML}{bcbd22}
\definecolor{python10}{HTML}{17becf}
\definecolor{designcolor}{HTML}{1f77b4}
\newcommand{\norm} [2][]{\left\|#2\right\|_{#1}}

\newtheorem{theorem}{Theorem}

\DeclareMathOperator*{\argmin}{argmin}

\def\mydefb#1{\expandafter\def\csname bb#1\endcsname{\mathbb{#1}}}
\def\mydefallb#1{\ifx#1\mydefallb\else\mydefb#1\expandafter\mydefallb\fi}
\mydefallb aAbBcCdDeEfFgGhHiIjJkKlLmMnNoOpPqQrRsStTuUvVwWxXyYzZ\mydefallb

\title{Variational Sparse Paired Autoencoders (vsPAIR) for Inverse Problems and Uncertainty Quantification}

\author[1]{Jack Michael Solomon$^{\dagger,}$}
\author[1]{Rishi Leburu}
\author[1]{Matthias Chung}
\affil[1]{Emory University Department of Mathematics}

\date{\today}

\begin{document}

\maketitle

\begin{abstract}
Inverse problems are fundamental to many scientific and engineering disciplines; they arise when one seeks to reconstruct hidden, underlying quantities from noisy measurements. Many applications demand not just point estimates but interpretable uncertainty. Providing fast inference alongside uncertainty estimates remains challenging yet desirable in numerous applications.  

We propose the Variational Sparse Paired Autoencoder (vsPAIR) to address this challenge. The architecture pairs a standard VAE encoding observations with a sparse VAE encoding quantities of interest (QoI), connected through a learned latent mapping. The variational structure enables uncertainty aware estimation, the paired architecture encourages interpretability by anchoring QoI representations to clean data, and sparse encodings provide structure by concentrating information into identifiable factors rather than diffusing across all dimensions. To validate the effectiveness of our proposed architecture, we conduct experiments on blind inpainting, computed tomography (CT), and initial-condition inference for the heat equation, demonstrating that vsPAIR is a capable inverse problem solver that can provide interpretable and structured uncertainty estimates.
\end{abstract}


\begingroup
\renewcommand\thefootnote{}
\footnotetext{${}^\dagger$: Corresponding author, direct questions to {\tt jack.michael.solomon@emory.edu}.}
\endgroup

\section{Introduction}\label{sec:introduction}

Inverse problems are central to scientific and engineering inference: from indirect and noisy measurements, one seeks to reconstruct hidden quantities that generated the data. Let $\mathcal{X}$ denote the space of unknowns (e.g., parameters or fields) and $\mathcal{Y}$ the space of observables, connected through a forward operator $F:\mathcal{X} \to \mathcal{Y}$. Measurements are modeled as
\[
    y = F(x) + \varepsilon,
\]
where $\varepsilon$ represents measurement noise or model error taking values in $\mathcal{Y}$.  
Because small perturbations in $y$ may induce large changes in the admissible $x$, such problems are typically \emph{ill-posed} in the sense of Hadamard \cite{hadamard1923lectures}.

Classical approaches address ill-posedness by introducing regularization or statistical priors. Variational methods such as Tikhonov, total variation, and sparsity-promoting penalties \cite{vogel2002computational,hansen2010discrete} stabilize the solution through explicit constraints, while Bayesian formulations interpret both $x$ and $y$ as random variables and infer the posterior $p(x \mid y)$ via Bayes' theorem \cite{calvetti2007introduction}. Although these approaches provide strong theoretical guarantees, they require repeated evaluation of $F$ and careful tuning of regularization parameters, which becomes computationally demanding in high-dimensional or nonlinear settings.

More recently, there has been a shift towards \emph{data-driven} approaches. Such approaches use parameterized neural networks to learn various attributes of the underlying inverse problem. \emph{End-to-end} methods learn direct mappings $F^{\gets}: \mathcal{Y} \rightarrow \mathcal{X}$ from paired samples $(x, y)$ \cite{jin2017deep, zhu2018image}. \emph{Prior learning} methods employ a data-driven prior: plug-and-play methods incorporate pretrained denoisers as implicit priors within optimization schemes \cite{venkatakrishnan2013plug, romano2017little}, and learned methods use regularization techniques to replace hand-crafted penalties with trainable functionals \cite{EHaber_2003, lunz2019adversarialregularizersinverseproblems}. Hybrid methods such as algorithm unrolling interpret iterative solvers as neural networks with learnable parameters \cite{Adler_2018, monga2021algorithm}. Various other methodologies exist and have been successful in a variety of applications \cite{chung2025good}.  

Learning latent representations of data offers a complementary regularization strategy. Rather than prescribing a hand-crafted regularizer, one learns low-complexity latent representations directly from data and performs inversion in this learned space. \emph{Autoencoders} and their probabilistic extensions, \emph{variational autoencoders (VAEs)} \cite{kingma2013auto}, capture intrinsic manifolds of high-dimensional data and thus serve as implicit priors. Standard autoencoders impose a low-dimensional bottleneck on the deterministic latent representation which typically has the effect of reducing high frequency variations \cite{hinton2006reducing, JMLR:v11:vincent10a}. Early works leveraged this regularizing property of under-complete autoencoders for applications in denoising \cite{JMLR:v11:vincent10a, bengio2013generalizeddenoisingautoencodersgenerative}. This later inspired the use of  autoencoders as an end-to-end approach \cite{9534012}. The implicit regularizing properties of autoencoders have also made them attractive for prior-learning where the learned distribution over the latent space captures the variability of plausible solutions and serves as a low-dimensional manifold constraint for reconstruction \cite{li2019nettsolvinginverseproblems}. Recent work on \emph{paired autoencoders}~\cite{chung2024paired,hart2025paired,chung2025good} and \emph{latent-twin architectures}~\cite{Chung_2026} extends this idea by coupling encoders and decoders in both $\mathcal{X}$- and $\mathcal{Y}$-domains, learning mappings between latent spaces that emulate both the forward and inverse operators. However, dense latent codes can entangle information, obscure uncertainty, and overfit small datasets.

A major drawback of many of the referenced approaches above is the lack of uncertainty quantification (UQ). In a Bayesian sense, UQ approximates the posterior $p(x \mid y)$, not merely a point estimate \cite{calvetti2007introduction}. Many applications demand this: in medical imaging, clinicians must distinguish pathology from reconstruction artifacts; in seismic inversion, uncertainty informs drilling risk; in autonomous systems, calibrated confidence is safety-critical. Score-based diffusion methods address UQ through iterative posterior sampling \cite{chung2023diffusion, song2021solving, kawar2022denoising}. Sample diversity reflects posterior uncertainty, but repeated network evaluations at inference sacrifice the speed of amortized approaches. The deterministic methods discussed above, including paired autoencoders, offer fast inference but lack the probabilistic structure necessary for UQ.

This motivates frameworks combining amortized inference with built-in probabilistic structure for { uncertainty aware estimates}. Sparse latent representations that correspond to interpretable features would additionally clarify which data characteristics drive reconstruction, and allow variance in estimates to be traced back to identifiable sources.

\paragraph{Contributions.} We propose the Variational Sparse Paired Autoencoder (vsPAIR), a framework combining probabilistic inference, sparsity and paired latent mappings for uncertainty-aware inversion. The architecture encodes observations $y$ via a standard VAE and sparse representations of quantities of interest $x$ via a \emph{sparse VAE} (sVAE), with a learned mapping connecting the two latent spaces. The sparse encoding reflects the assumption that observations are driven by few dominant factors, yielding interpretable latent codes while regularizing reconstruction. The variational structure enables { approximate UQ}.

The key contributions of this work are as follows:
\begin{enumerate}
    \item The vsPAIR framework with probabilistic latent structure, enabling both fast inference and { explainable variance in reconstructions}.
    \item Experimental validation on ill-posed inverse problems--blind inpainting, computed tomography (CT) { and nonlinear heat equation initial condition inference--that demonstrate competitive reconstruction performance}.
    \item { Demonstration that vsPAIR yields sparse, interpretable latent representations in which active dimensions correspond to core reconstruction features and their associated uncertainty localizes to those features.}
\end{enumerate}

\paragraph{Structure of the paper.}
\Cref{sec:background} reviews classical and learning-based formulations of inverse problems, including autoencoders, paired latent mappings { and generative approaches to UQ}. \Cref{sec:method} introduces the vsPAIR framework. Finally, \Cref{sec:numeric} and \Cref{sec:conclusion} present numerical demonstrations on blind inpainting CT, and heat inference, followed by concluding remarks. { The introduction of sparsity in the QoI autoencoder presents technical challenges which are discussed and addressed in \Cref{appendix:svae}}.

\paragraph{Notation and Setting.} Let $X$ denote the quantity of interest (QoI) and $Y$ the observation, interpreted as random variables with a joint distribution $(X, Y)$ absolutely continuous with respect to Lebesgue measure, thus admitting a joint density $p(x,y)$ \cite{DU04}. We use $p$ for true distributions and $q$ for variational approximations. Notation is summarized in \Cref{tab:notation}.

\begin{table}[h]
\centering
\begin{tabular}{@{}cl@{}}
\toprule
Symbol & Description \\
\midrule
$X, Y, Z$ & Random variables for QoI, observation, and latent codes \\
$x, y, z$ & Samples from respective distributions \\
$p(\cdot)$, $q(\cdot)$ & True and approximate (variational) distributions\\
$q_{\phi_x}(z \mid x)$ & Variational encoder for QoI with trainable parameters $\phi_x$ \\
$q_{\theta_x}(x \mid z)$ & Probabilistic decoder for QoI with trainable parameters $\theta_x$ \\
$q_{\phi_y}(z \mid y)$, & Variational encoder for observations with trainable parameters $\phi_y$\\
 $q_{\theta_y}(y \mid z)$ & Probabilistic decoder for observations with trainable parameters, $\theta_y$ \\
$\mu_{\phi_x}, \sigma_{\phi_x}, \omega_{\phi_x}$ & Network outputs for that parameterize the spike-and-slab posterior \\
$\mu_{\phi_y}, \sigma_{\phi_y}$ & Network outputs for that parameterize the Gaussian posterior \\
$z_x, z_y$ & Latent samples: $z_x \sim q_{\phi_x}(z \mid x)$, $z_y \sim q_{\phi_y}(z \mid y)$ \\
$M^{\gets}_{\theta_M}$ & Learned mapping $\mathcal{Z}_Y \to \mathcal{Z}_X$ with trainable parameters $\theta_M$ \\
$\tilde{x}, \tilde{y}$ & Autoencoder reconstructions of $x$, $y$ \\
$\hat{x}$, $\hat{z_x}$ & Reconstructions of QoI and its latent representation \\
$\mathcal{N}(v \mid \mu, \Sigma)$ & Normal distribution with mean $\mu$ and covariance $\Sigma$\\
$\mathcal{SS}(v \mid \mu, \sigma, \omega)$ & Spike-and-slab distribution with parameters $\mu, \sigma$ and $\omega$
\\
\bottomrule
\end{tabular}
\caption{Summary of notation.}
\label{tab:notation}
\end{table}

\section{Background}\label{sec:background}

This section reviews the classical formulations of inverse problems and the learning-based methods that motivate our approach. { The reader is referred to \cite{chung2025good} for a more in-depth exposition; we closely follow the structure and notation of this work. }

Inverse problems seek to recover unknown quantities from indirect observations. We consider the forward operator $F:\mathcal{X}\to\mathcal{Y}$, which maps unknowns to observables. In many applications, this operator naturally decomposes as
\[
    F = P \circ u,
\]
where $u:\mathcal{X}\to\mathcal{U}$ maps model parameters $x\in\mathcal{X}$ to a state $u(x)\in\mathcal{U}$ given by the solution of a governing ordinary or partial differential equation (ODE/PDE), and $P:\mathcal{U}\to\mathcal{Y}$ projects the state onto the observation space. Observations then satisfy
\begin{align}\label{eq:fwdprocess}
    y = F(x) + \varepsilon = (P\circ u)(x) + \varepsilon,
\end{align}
with noise or model discrepancy $\varepsilon\in\mathcal{Y}$. This decomposition clarifies where computational cost concentrates, namely in solving for $u$, and where structural priors are specified on $x$ and/or $u$.

A deterministic estimate $\hat x$ is commonly obtained by minimizing a stabilized objective
\begin{equation}\label{eq:varinverse}
   \hat{x} =  \argmin_{x\in\mathcal{X}} \
    \mathcal{C} \big(F(x),y\big) + \mathcal{R}(x),
\end{equation}
where $\mathcal{C}$ measures data fidelity (e.g., $\|F(x)-y\|_{2}^{2}$ or a negative log-likelihood) and $\mathcal{R}$ encodes prior structure. Canonical choices for $\mathcal{R}$ include quadratic Tikhonov, total-variation, and $\ell^1$ sparsity penalties \cite{vogel2002computational, hansen2010discrete}. These restore stability but require careful selection of weights and may still be expensive when $F$ is nonlinear and high-dimensional.

The Bayesian perspective specifies a prior $p_{\rm prior}(x)$ and likelihood $p_{\rm like}(y \mid x)$, yielding the posterior
\[
    p_{\rm post}(x\mid y) \propto  p_{\rm like}(y\mid x)\,p_{\rm prior}(x),
\]
which quantifies uncertainty in the unknown $x$, given observation $y$ and prior knowledge on $x$, \cite{stuart2010inverse, calvetti2007introduction}. Under a Gaussian noise assumption $\varepsilon\sim\mathcal{N}(\varepsilon \mid 0,\Gamma_y)$ and Gaussian prior $x\sim\mathcal{N}(x \mid m,\Gamma_x)$ with with $\Gamma_x, \Gamma_y$, symmetric positive definite, the maximum a posteriori (MAP) estimator solves
\begin{equation}\label{eq:MAP}
    \min_{x\in\mathcal{X}} \
    \tfrac12\big\|F(x)-y\big\|_{\Gamma_y^{-1}}^{2} + \tfrac12\big\|x-m\big\|_{\Gamma_x^{-1}}^{2},
\end{equation}
which recovers the Tikhonov problem \Cref{eq:varinverse}. Thus, quadratic variational regularization corresponds to a Gaussian prior; more generally, the negative log-prior acts as a regularizer $\mathcal{R}$.

When one expects the unknown $x$ to be compressible, an $\ell^1$ penalty in \Cref{eq:varinverse} corresponds to a Laplace prior, promoting sparse solutions while retaining convexity in many linear cases \cite{candes2008restricted}. Such sparsity assumptions are common in imaging and geophysical inversion, and motivate the latent sparsity we later enforce in our learned surrogates \cite{chung2023variable, solomon2025fast}. 

Both paradigms face scalability limits. Solving \Cref{eq:varinverse} for a point-estimate via gradient-based methods typically requires repeated evaluations of $F$ and its derivatives; sampling or approximating $p_{\rm post}$ may become prohibitive for nonlinear $F$ or large $\dim(\mathcal{X})$. The intractability of these solvers is further amplified for UQ.  Iterative approaches such as Markov Chain Monte Carlo and Kalman Filtration require repeated computations using the forward model, which quickly render them infeasible for use in large problems without simplification or approximation \cite{stuart2010inverse}.  Moreover, these approaches are inaccessible when $F$ is unknown or unavailable. 

These costs motivate surrogate models that reduce the expenses associated with inverse inference.  Many of these methods are based around dimensionality reduction which formulate the problem in a lower dimensional space.  In addition to the reducing computational costs, such approaches are attractive since the dimensionality bottleneck induces an implicit prior.  Recent years have seen a surge in scientific machine learning approaches that follow this paradigm. Deep neural networks can approximate complex operators in high dimensions \cite{HORNIK1989359}. Autoencoders and VAEs \cite{kingma2013auto} embrace a low-dimensional latent space as their core feature, using self-supervised learning to generate compressed representations of data.  The defining feature of latent-twins approaches \cite{Chung_2026} is to learn the inversion between latent representations, which is embraced in the vsPAIR framework.  

The following section (\Cref{subsec:autoreview}) provides background on autoencoders, VAEs and sparse VAEs (sVAE), a description of paired autoencoders (\Cref{subsec:pair}) concluded with a discussion on previous data-driven { and generative} approaches to UQ (\Cref{subsec:prevwork}).

\subsection{Variational Autoencoders (VAEs) and Sparse Variational Autoencoders (sVAEs)} \label{subsec:autoreview}
Autoencoders have become a cornerstone of machine learning and have seen significant application to inverse problems. These models learn a parameterized mapping from the data space to itself via a latent representation. Let $\mathcal{X} \subset \bbR^{n}$ be the data space. An autoencoder consists of two components: a parameterized \emph{encoder} $e_{\phi}: \mathcal{X} \rightarrow \mathcal{Z}$ mapping data to a latent space $\mathcal{Z} \subset \bbR^{\ell}$, and a parameterized \emph{decoder} $d_{\theta}: \mathcal{Z} \rightarrow \mathcal{X}$ mapping back to the data space, where $\phi$ and $\theta$ are corresponding trainable network parameters. Together, these networks are trained such that $\tilde{x}=d_{\theta}\circ e_{\phi}(x) \approx x$. When $\ell < n$, the dimensional bottleneck encourages the model to learn compact, meaningful representations. Autoencoders are broadly classified into \emph{undercomplete} ($\ell < n$) and \emph{overcomplete} ($\ell > n$).  Overcomplete autoencoders can also learn compressed representations of data by enforcing sparsity in the latent representation \cite{Chung_2025}. For a more extensive review on autoencoders, see \cite{chung2025good}.

\vspace{1em}
\emph{Variational Autoencoders} (VAEs)~\cite{kingma2013auto} extend traditional autoencoders into the probabilistic domain, enabling generative modeling of data. Instead of deterministic mappings, VAEs learn a parameterized stochastic encoder or \emph{variational posterior} $q_\phi(z \mid x)$ approximating the posterior over latent variables, and a parameterized probabilistic decoder or \emph{likelihood model} $q_\theta(x \mid z)$ reconstructing data from latent samples. Deep Neural Networks with trainable parameters $\phi$ and $\theta$ are used to learn the a mapping from $x$ to the distribution parameters of $q_\phi(z \mid x)$ and $z$ to the distribution parameters of $q_\theta(x \mid z)$, respectively.  Training maximizes the evidence lower bound (ELBO), a lower bound on the log-likelihood of the generated sample:
\begin{align}
\label{eq:elbo}
\log p(x) \geq \underbrace{\mathbb{E}_{z \sim q_\phi(z \mid x)} \big[ \log q_\theta(x \mid z) \big]}_{\text{reconstruction}}
- \underbrace{{\rm D}_{\mathrm{KL}}\!\big( q_\phi(z \mid x) \; || \; p(z) \big)}_{\text{regularization}},
\end{align}
where $p(z)$ is an assumed prior over the latent space, typically $\mathcal{N}(z \mid 0, I)$, and ${\rm D}_{\rm KL}(\, \cdot \; || \; \cdot\, )$ denotes the KL-divergence \cite{pml1Book}. Under Gaussian assumptions on the probabilistic encoder and decoder as well as unit Gaussian assumption on the prior, the reconstruction term reduces to scaled mean-squared error and the KL term admits a closed form; see \cite{kingma2019introduction} for a comprehensive tutorial. For further simplicity, we also assume that the encoder and decoder have diagonal and unit covariances, respectively. Negating \Cref{eq:elbo}, and plugging in these expressions, we obtain the VAE objective to be minimized in training:
\begin{align}\label{eq:fullvaeloss}
\mathcal{L}_{\rm VAE}(\phi, \theta) = \frac{1}{2} \, \mathbb{E}_{z \sim q_\phi(z \mid x)} \big[ \| x - \mu_\theta(z) \|^2 \big] +  \frac{1}{2} \sum_{j=1}^\ell \left( \sigma^2_{\phi,j}(x) + \mu^2_{\phi,j}(x) - 1 - \log \sigma^2_{\phi,j}(x) \right),
\end{align}
where $\mu_\phi(x)$ $\sigma_\phi(x)$ are learned neural networks to compute the mean and diagonal covariance of the encoder distribution: $q_\phi(z \mid x) = \mathcal{N}(z \mid \mu_\phi(x), {\rm diag} (\sigma^2_\phi(x)))$; $\mu_\theta(z)$ is the learned network to compute the mean of the decoder distribution: $q_\theta(x \mid z) = \mathcal{N}(x \mid \mu_\theta(z), I )$.  The ${\rm diag}(\cdot)$ will be omitted for readability.

The KL regularization encourages the learned posteriors to match the prior, enabling generation by sampling $z \sim p(z)$ and decoding. However, this regularization introduces a fundamental tradeoff: pushing $q_\phi(z \mid x)$ toward $p(z)$ reduces the mutual information between $x$ and $z$, limiting reconstruction fidelity \cite{alemi2018fixing}. This tension between reconstruction quality and generative structure is inherent to the VAE framework.  To make stochastic sampling from the latent space compatible with gradient-based optimization, VAEs employ the \emph{reparameterization trick}, see \cite{kingma2013auto}.

Typical VAEs will utilize all of the latent dimensions regardless of whether they hold relevant information or not.  Variational Sparse Autoencoders (sVAEs) extend the classical VAE framework by explicitly encouraging sparse latent representations. 

The sVAE framework was introduced in \cite{tonoliniVariationalSparseCoding2020}, where the authors proposed using a \emph{spike-and-slab distribution} for both the prior and variational posterior and derived the corresponding closed-form KL divergence enabling ELBO-based training. { A spike-and-slab distribution is a mixture of a \emph{spike} or point mass and continuous \emph{slab} distribution} often taking the form of a multivariate Gaussian with mixture weight $\omega$ \cite{mitchell1988bayesian}.  The distribution can be sampled according to the following procedure:
\begin{gather}
        \nu \sim \text{Bernoulli}(\omega) \label{ln:om1}\\ 
            z\mid \nu = 0 \sim \delta_0(z) \quad \text{ and } \quad z \mid \nu= 1 \sim \mathcal{N}(z \mid \mu,\sigma^2). \label{ln:om2}
\end{gather}
In the context of variational encoding, if we assume that each {latent }index follows a spike-and-slab distribution with a Gaussian slab, i.e.,
\[
q_{\phi}(z_i  \mid x) =   \omega_i \mathcal{N}(z_i \mid \mu_i,\sigma_i^2) +  (1-\omega_i) \delta_0(z_i) 
\]
then there is a probability of $\omega_i$ of the $i$-th index being switched on, in which case it follows a Gaussian distribution with parameters $\mu_i$ and $\sigma_i$, and a $1-\omega_i$ probability of being switched off and set to 0.  A spike-and-slab distribution for the prior latent distribution is assumed similarly,
\begin{align}
p(z_i) =   \rho_i\mathcal{N}(z_i \mid {0, 1}) + (1-\rho_i)\delta_0(z_i),
\end{align}
where $\rho_i$ denotes the prior probability of the $i$-th index being active.  For simplicity, {we assume that this prior is fixed across the dataset, and that this probability is constant for each index, i.e., $\rho_i = \rho$.  The original implementation in \cite{tonoliniVariationalSparseCoding2020} leveraged a classifier-based approach to encourage input-dependent selectivity in the active indices.} Consistent with the standard VAE, the parameters for the spike-and-slab distribution of the encoder are given by neural network outputs, dependent on $x$: $\mu_\phi(x), \sigma_\phi(x), \omega_\phi(x)$.

We denote the spike-and-slab sampling process (\Cref{ln:om1} and \Cref{ln:om2}) as $z \sim \mathcal{SS}(z \mid \mu, \sigma, \omega)$. Note $p(z_i \mid x)$ is not a density with respect to the Lebesgue measure since the distribution assigns non-zero probability with respect to a single point, i.e., the measure is not absolutely continuous with respect to the Lebesgue measure.  This distribution should instead be interpreted as a mixture of a discrete spike and a continuous slab, where the expression above denotes the corresponding density with respect to an augmented Lebesgue measure. From the result in \cite{tonoliniVariationalSparseCoding2020}, when the slab component of both the encoder and prior are Gaussian { with spikes at 0}, {the KL divergence between two spike-and-slab distributions is over all $\ell$ latent dimensions is given by}
\begin{equation}\label{eq:ss_kl_full}
{\rm D}_{\mathrm{KL}}(q(z \mid x) \; ||\; p(z)) = \sum_{j=1}^\ell \left[  \frac{\omega_j}{2}\left( \sigma_j^2 + \mu_j^2 - 1 - \log\sigma_j^2 \right) + (1-\omega_j)\log\frac{1-\omega_j}{1-\rho} + \omega_j\log\frac{\omega_j}{\rho}  \right].
\end{equation}
Combined with a reconstruction term for a Gaussian decoder with unit covariance, the negative ELBO for the sVAE becomes
\begin{equation}\label{eq:svae_elbo}
\mathcal{L}_{\text{sVAE}}(\phi, \theta) = \frac{1}{2} \mathbb{E}_{z \sim q_\phi(z \mid x)}\bigl[\|x-\mu_\theta(z)\|^2 \bigr] + \sum_{j=1}^\ell \left[  \frac{\omega_j}{2}\left( \sigma_j^2 + \mu_j^2 - 1 - \log\sigma_j^2 \right) + (1-\omega_j)\log\frac{1-\omega_j}{1-\rho} + \omega_j\log\frac{\omega_j}{\rho}  \right].
\end{equation}

Training of the sVAE must be handled carefully--the immediate question of how to choose a prior sparsity level $\rho$ arises and the Bernoulli sampling procedure further introduces complexity due to non-differentiability.  These challenges are discussed in \Cref{appendix:svae}, { where we describe a training objective we use to learn $\rho$.}

The benefit of sVAEs are that they typically lead to more interpretable, disentangled latent representations \cite{tonoliniVariationalSparseCoding2020}. Because many real‐world signals lie on low‐dimensional manifolds, enforcing sparsity helps the model focus its capacity on the most meaningful latent factors, rather than spreading small amounts of information over all dimensions. They bridge the gap between classical sparse coding, which enforces strict zeros, and VAEs, which offer a probabilistic formulation but no mechanism for efficient sparse representation.

\subsection{Paired Autoencoder Frameworks (Latent Twins)}\label{subsec:pair} Introduced in \cite{Chung_2026}, the latent twin methodology learns mathematical operators between latent representations of quantities of interest.  This approach is broadly applicable to many areas of research including PDEs, ODEs, and inverse problems and can be adjusted to encode physical constraints on the system of interest in the learned latent space.  To be precise, our work builds off of paired autoencoders for inverse problems, a precursor to the latent twin architecture. { Once again, we follow the notation and formulation used in \cite{chung2025good}}.

\begin{figure}[htb!]
    \centering
    \includegraphics[width=0.95\textwidth]{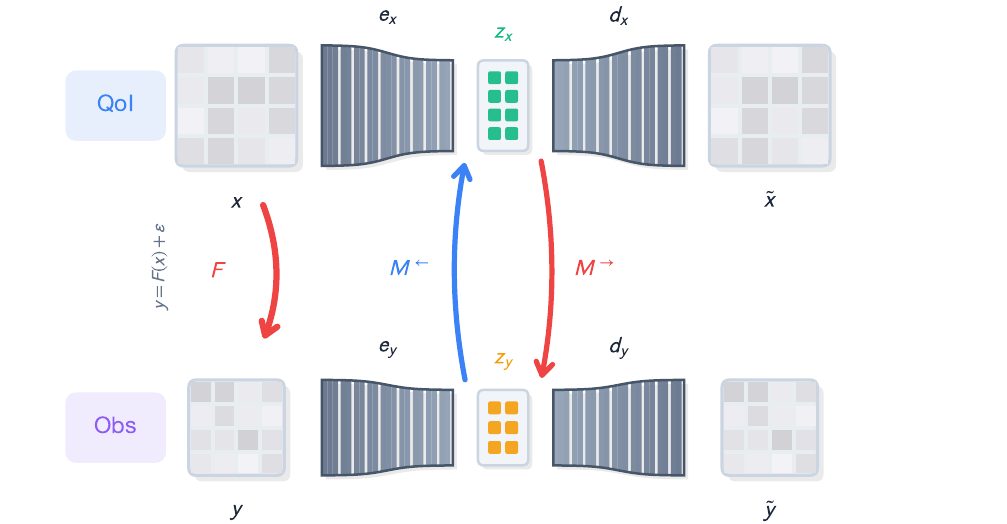}
    \caption{Deterministic paired autoencoder (PAIR) architecture where the QoI and observation are related by a forward process, $F$: $y = F(x) + \varepsilon$.  Separate Autoencoders are trained to encode the QoI ($x$) and observation ($y$), as well as surrogate forward $M^\to$ and inversion $M^\gets$ inversion operators that operate on the latent encoding.}
    \label{fig:pair}
\end{figure}

Paired autoencoder frameworks for inverse problems have been studied extensively in previous works \cite{boink2020learnedsvdsolvinginverse, piening2024pairedwassersteinautoencodersconditional, hart2025paired, chung2024paired}.  Here, two autoencoders are learned: one for the quantity of interest $x \in \mathcal{X}$ and one for the data $y \in \mathcal{Y}$: each comprises an encoder-decoder pair ($e_{\phi_x}$, $d_{\theta_x}$) and ($e_{\phi_y}$, $d_{\theta_y}$).  For ease of notation, we suppress the dependence of the encoders/decoders on trained parameters,  replacing the subscript with the variable the networks are trained for; the latent spaces of these autoencoders are denoted $\mathcal{Z}_x \subset \mathbb{R}^{\ell_x}$ and $\mathcal{Z}_y \subset \mathbb{R}^{\ell_y}$, respectively. The innovation lies in jointly training mappings $M^\to: \mathcal{Z}_x \to \mathcal{Z}_y$ and $M^\gets: \mathcal{Z}_y \to \mathcal{Z}_x$ to bridge these spaces, compare \Cref{fig:pair}. $M^{\to}, \text{and } M^\gets$, are both parameterized by its respective network parameters $ \theta_{M^\to}, \text{and } \theta_{M^\gets}$, which are also omitted unless directly relevant to discussion.  These compose surrogate forward and inversion operators that act on the latent spaces of the autoencoders.

The aim of surrogate inverse modeling is to provide a direct mapping from the observable $y$ to the unobservable quantity $x$. The resulting compositions $d_y \circ M^\to \circ e_x$ and $d_x \circ M^\gets \circ e_y$ provide surrogate forward 
\begin{equation}
    \hat{y} = (d_y \circ M^\to \circ e_x)(x) \quad \text{and inverse mapping} \qquad \hat{x} = (d_x \circ M^\gets \circ e_y)(y).
\end{equation}
Note that the hat indicates an inferred variable using either $M^{\to}$ or $M^{\gets}$ while the tilde indicates a direct pass through one of the autoencoders.  A paired autoencoder may be learned on training data provided from the corresponding distributions by minimizing a combination of reconstruction and cross-space surrogate losses
\begin{align}\label{eq:lossfcns}
    &\mathbb{E}_{x \sim p(x)} \ \mathcal{J}((d_x \circ e_x)(x),x), \quad \mathbb{E}_{y \sim p(y)} \ \mathcal{J}((d_y \circ e_y)(y),y), \\
    &\mathbb{E}_{x \sim p(x)} \ \mathcal{J}(d_y \circ M^\to \circ e_x (x), y), \quad \mathbb{E}_{y \sim p(y)} \ \mathcal{J}(d_x \circ M^\gets \circ e_y (y), x).
\end{align}
where $\mathcal{J}$ are appropriate losses, i.e., mean-squared-error (MSE).  A design choice must be made as to whether the autoencoders are trained independently, or if the whole framework is trained together; the latter typically demonstrates better performance.  For details on training paired autoencoders the interested reader is referred to \cite{chung2024paired,chung2025good}. In particular, our work builds off of the following innovations: \cite{chung2024paired} introduced the paired-autoencoder (PAIR) architecture that our framework is based on; \cite{piening2024pairedwassersteinautoencodersconditional} is a generative adaptation of the approach which instead uses Wasserstein Autoencoders \cite{tolstikhin2019wassersteinautoencoders}; \cite{Chung_2025} demonstrated the benefits of sparse overcomplete autoencoders for learning sparse embeddings of data and { \cite{doi:10.1021/jacsau.3c00275} introduces a similar paired VAE architecture for material science applications that couples two VAEs by aligning their latent spaces}.  These methods have been successfully applied across a wide range of problems.

{\subsection{Generative Approaches for Inverse Problems}\label{subsec:prevwork}
{

Prior to detailing the vsPAIR approach, we provide background on related approaches. The advent of machine learning has led to a variety of data-driven approaches for inverse problems and UQ.  Many data-driven strategies have been successfully applied to pointwise inversion, such as the \emph{end-to-end}, \emph{prior learning} and \emph{algorithm unrolling} strategies discussed in \Cref{sec:introduction}. Generally, these methods provide high fidelity point estimates but are not well suited for application in UQ.

Recent advances in generative modeling have seen the development of numerous principled approaches to UQ in inverse problems.  These methodologies generally seek to estimate the posterior $p(x \mid y)$ using a learned approximate posterior  and have exploited the properties of many generative frameworks.

VAEs have been used extensively in this area.  Conditional VAEs (CVAE) \cite{NIPS2015_8d55a249} model $p(x \mid y)$ by conditioning the latent variable on the observation, $y$.  Other approaches use the forward model to link latent variables with the QoI. For example, the approach in \cite{goh2021solvingbayesianinverseproblems} identifies a VAE latent space with the QoI and the decoder with the forward model so that the encoder approximates the posterior. In \cite{zhang2021conditionalvariationalautoencoderlearned}, the authors develop a CVAE framework for image reconstruction, conditioning the latent variable on the observation and incorporating the forward operator into a recurrent reconstruction network to shape reconstructions.

Other strategies use Normalizing Flows--a family of generative models in which a complex distribution is approximated by the learned pushforward of a simple distribution through an invertible map \cite{Tomczak2024}.  The approach in \cite{siahkoohi2020fasteruncertaintyquantificationinverse} leverages Conditional Normalizing Flows (CNF) to approximate the posterior as the push-forward of a simple base distribution through an observation-aware invertible map. This is then used to accelerate inference on related unsupervised problems. In \cite{pmlr-v139-whang21b} a pretrained unconditional normalizing flow is used as a prior on the QoI and the authors establish a variational framework for approximate conditional inference. In the context of CT, \cite{denker2020conditionalnormalizingflowslowdose} uses a CNF informed by the forward process through a filtered back projection conditioner
to approximate the posterior, demonstrating high fidelity reconstructions alongside pointwise UQ estimates. An alternative approach is taken in  \cite{sun2020deepprobabilisticimaginguncertainty} where the authors focus on estimates for a single observation.  They leverage an untrained flow model to approximate the image posterior for characterization of multi-modal reconstructions. 

Diffusion and score-based models \cite{song2021scorebasedgenerativemodelingstochastic} have emerged as a powerful class of generative models to represent prior distributions for inverse problems.  These methods start with the QoI and iteratively add noise until the result is approximately Gaussian. A neural network is used to estimate the score of the corrupted image.  The learned score is then used to reverse the diffusion process: prior samples are generated by sampling a unit Gaussian and iteratively denoising using the learned process.  In \cite{song2021solving}, the authors apply score models to medical inverse problems, training a score network for the prior distribution of the QoI, and augment sampling to enforce consistency with observations using the forward model.  Similarly, \cite{kawar2022denoising} proposes DDRM which leverages a pre-trained diffusion model for the QoI prior and uses the Singular Value Decomposition of the forward model to inform the reverse diffusion process. 
Diffusion Posterior Sampling (DPS) \cite{chung2023diffusion} offers a framework for approximating posterior sampling, suitable for nonlinear inverse problems.  Like other methods, it starts with a score network learned as the unconditional prior for the QoI.  Given an observation, the posterior score is decomposed into the prior score and likelihood term, the former given by the score network and latter approximated.  This results in a tractable gradient term to inform the diffusion process for approximate posterior sampling.  

The key advantage of many of these generative approaches is that they provide not only viable pointwise reconstructions, but approximate posterior sampling.  This enables characterization of the posterior distribution through repeated sampling.  We use DPS as a baseline to benchmark the UQ properties of the vsPAIR approach. 

Like many of the referenced works, vsPAIR seeks to supplement pointwise estimates of the QoI with meaningful uncertainty information.  vsPAIR diverges from the works referenced in this section in two key ways--first, vsPAIR offers the advantage that specification of the forward process is not required.  Only QoI, observation pairs are required for training.  Second, vsPAIR seeks structured uncertainty.  Existing methods characterize uncertainty through dense representations of the posterior.  vsPAIR introduces sparsity to concentrate uncertainty information into a sparse set of indices. However, the tradeoff is that vsPAIR does not target UQ estimates that are calibrated with the posterior.  Instead, reported uncertainties reflect where the model detects variation in plausible reconstructions, often localized to identifiable features.

}}

\section{Methodology: vsPAIR Framework}\label{sec:method}

In this section, we present the key contribution of this work: the Variational Sparse Paired Autoencoder (vsPAIR). { For details on the sVAE implementation, see \Cref{appendix:svae}}.

\begin{figure}[htb!]
    \centering
    \includegraphics[width=1.0\textwidth]{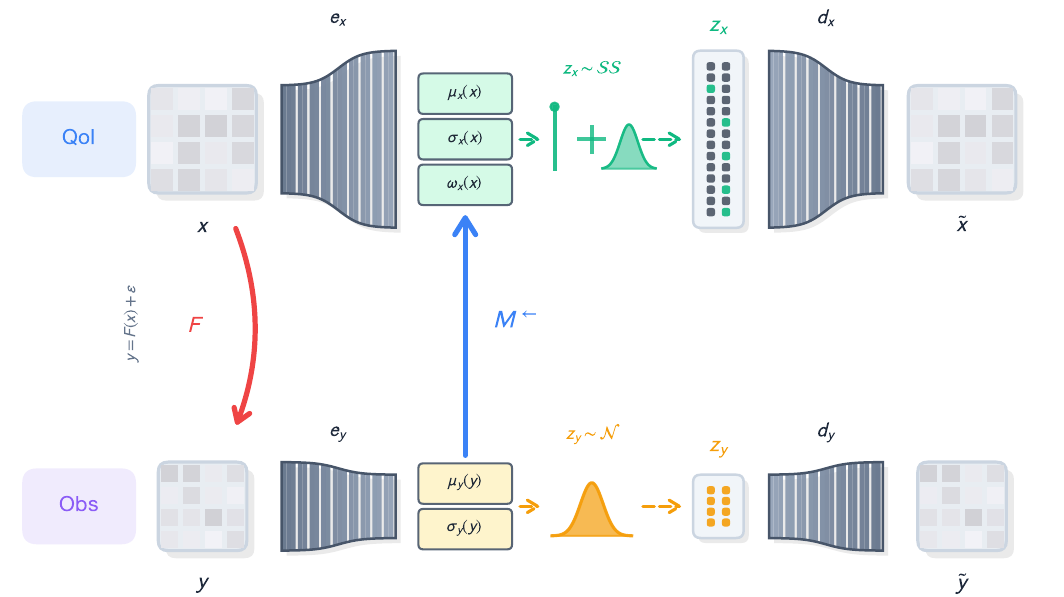}
    \caption{Schematic of the vsPAIR framework. The QoI is encoded via an sVAE producing sparse latent representations $z_x$, while observations are encoded via a standard VAE. The latent mapping $M^\gets$ operates on distribution parameters, mapping $(\mu_y, \sigma_y)$ to $(\hat{\mu}_x, \hat{\sigma}_x, \hat{\omega}_x)$.}
    \label{fig:vspair_arch}
\end{figure}

The vsPAIR framework pairs a standard VAE encoding observations with a sVAE encoding QoI, connected through a learned latent mapping that serves as a surrogate inversion operator. Specifically, observations are encoded via the Gaussian variational posterior $z_y \sim q_{\phi_y}(z_y \mid y)$, while the QoI is encoded using the sVAE spike-and-slab posterior $z_x \sim q_{\phi_x}(z_x \mid x)$. A trainable latent mapping $M^\gets_{\theta_M}$ connects these two representations, mapping from the observation latent space to the QoI latent space. $\theta_{M}$ denotes the trainable parameters of this mapping and are suppressed for ease of notation.  
Inversion of an observation proceeds by encoding $y$ to $z_y$, using $M^{\gets}$  to obtain a representation $\hat{z}_x \in \mathcal{Z}_x$, and then decoding the representation to generate a reconstruction: $\hat{x}$. A schematic of the architecture is shown in \Cref{fig:vspair_arch}.

Training follows the structure of the deterministic PAIR framework \cite{hart2025paired}, with reconstruction losses replaced by the corresponding ELBOs. The complete objective is
\begin{align}\label{eq:vspair_loss}
    \mathcal{L}(\phi_{x}, \theta_{x}, \phi_{y}, \theta_{y}, \theta_{M}, \rho)  = \mathbb{E}_{(x, y) \sim p(x, y)} \big[\lambda_1 \mathcal{L}_{\text{sVAE}}(\phi_{x}, \theta_{x}) + \lambda_2 \mathcal{L}_{\text{VAE}}(\phi_{y}, \theta_{y}) + \lambda_3 \mathcal{L}_{M}(\theta_M)  + \lambda_\rho \mathcal{L}_\rho(\rho\, ; \, \alpha_0, \beta_0)\big],
\end{align}
where $\mathcal{L}_{\text{sVAE}}$ (\Cref{eq:svae_elbo}) and $\mathcal{L}_{\text{VAE}}$ (\Cref{eq:fullvaeloss}) are the ELBO losses for the QoI and observation autoencoders respectively, $\mathcal{L}_{M}$ is the latent mapping loss defined below and  { $\mathcal{L}(\rho \, ; \, \alpha_0, \beta_0)$ is a Beta regularization term used to learn the prior sparsity level--see \Cref{appendix:svae}, \Cref{eq:svae_loss_full}.} The non-negative hyperparameters $\lambda_1, \ldots, \lambda_\rho$ control the relative weighting of each term. {The expectation denotes averaging over labeled data pairs $(x, y)$ sampled from the data distribution $p(x, y)$ prior to training.}

{ The core application of the vsPAIR framework that we explore is inverse problems.  In alignment with \Cref{eq:fwdprocess}, we assume that observations, $y$ are statistically related to $x$ via the forward process:
\begin{align}
    y = F(x) + \varepsilon.
\end{align}
The vsPAIR encoding process can be represented as a graphical network with nodes corresponding to $x$, $y$, $z_x$ and $z_y$--see \Cref{fig:Bayesnet}.
\begin{figure}
    \centering
    \begin{tikzpicture}[
        latent/.style={
            circle, draw=black!70, thick,
            minimum size=1cm, inner sep=2pt,
            font=\large
        },
        observed/.style={
            circle, draw=black!70, thick, fill=gray!15,
            minimum size=1cm, inner sep=2pt,
            font=\large
        },
        arr/.style={
            ->, thick, draw=black!60
        }
    ]
     
    \node[observed] (x)  at (0,0)   {$x$};
    \node[observed] (y)  at (2,0)   {$y$};
    \node[latent]   (zx) at (0,-2)  {$z_x$};
    \node[latent]   (zy) at (2,-2)  {$z_y$};
     
    \draw[arr] (x) -- (zx);
    \draw[arr] (x) -- (y);
    \draw[arr] (y) -- (zy);
     
    \end{tikzpicture} 
    \caption{Graphical representation underlying the structure of vsPAIR encoding process.  Shaded nodes are observed during training, and unshaded correspond to latent encodings.  $y$ is generated from $x$ through the forward model while latent variables are obtained by encoding their parent node. We  assume deterministic decoders throughout this work.}
    \label{fig:Bayesnet}
\end{figure}
The paired structure assumes the factorization $p(x, y, z_x, z_y) = p(x)p(y \mid x) p(z_x \mid x) p(z_y \mid y)$ for the encoding process which implies following conditional independencies in the latent variables: $z_x$ is conditionally independent of $y$ given $x$ and $z_y$ is conditionally independent of $x$ given $y$. In the vsPAIR formulation, each latent variable is assigned a prior in alignment with the type of VAE used: $z_x$ is assigned a spike-and-slab prior, while $z_y$ is assigned a standard Gaussian prior.  These priors enter the objective through the respective ELBO terms.  The encoder distributions, $q_{\phi_x}(z_x \mid x)$ and $q_{\phi_y}(z_y \mid y)$ serve as variational approximations to the respective posteriors. }

{ The first design choice that must be made is whether the autoencoders are to be trained first, and then the latent map after, or if training {should} proceed all together. If the autoencoders are trained independently, then each encoder learns network parameters from their respective inputs alone.  In the case that they are trained jointly with the full objective, network parameters of each respective encoder may be influenced by the relationship between $x$ and $y$ and capture richer cross-variable dependencies that independent training cannot capture.  The implications  of this choice were investigated for the deterministic case in \cite{hart2025paired}, and empirically, coupled training of $M^\gets$ yielded better results. In either case, the graph is unchanged as each encoder only takes in $x$ or $y$ to generate $z_x$ or $z_y$ respectively.  This independence is an architectural property assumed by the paired framework, not an assumption about the true data distribution.}

The second design choice concerns the implementation of the latent mapping. For simplicity, we once again omit dependence on trainable parameters, that is, $\mu_{\phi_x}(x)$ is written as $\mu_x(x)$. We propose learning $M^{\gets}$ as a direct mapping from the distribution parameters of the observation encoder to the distribution parameters of the QoI encoder: $M^\gets$ learns to map $(\mu_{y}(y), \sigma_{y}(y))$ to $(\mu_x(x), \sigma_x(x), \omega_{x}(x))$ for $(x, y)\sim p(x, y)$.  Here, inversion proceeds by first encoding the observation $y$, then computing approximate conditional distribution parameters for $x$,   $(\hat{\mu}(y), \hat\sigma(y), \hat\omega(y))$, via $M^\gets$, then decoding a sample $\hat{z}_x$ from this approximate distribution to obtain $\hat{x}$.  Generation can be accomplished by sampling the approximate latent distribution and decoding. $M^{\gets}$ can be trained via MSE loss between the true and approximate parameters:
\begin{align}
\mathcal{L}_M(\theta_M) = \norm[2]{\begin{bmatrix}
    \hat{\mu}_{\theta_{M}}(y)\\
    \hat{\Sigma}_{\theta_{M}}(y) \\
    \hat{\omega}_{\theta_{M}}(y)
\end{bmatrix} -  \begin{bmatrix}
    \mu_{x}(x) \\ 
    \Sigma_{x}(x) \\ 
    \omega_{x}(x)
\end{bmatrix}}^2 + \lambda_b \sum_{j=1}^{\ell} \hat{\omega}_j(y)(1 - \hat{\omega}_j(y)). 
\end{align}  
In order to prevent excessive averaging in the $\omega$ parameter estimation, we append an optional second term: a binary entropy penalty that encourages decisive gating predictions.  This term equals zero when $\hat{\omega}_j \in \{0, 1\}$ and is maximized at $\hat{\omega}_j = 0.5$, pushing the mapping to commit to activation or deactivation rather than hedging with intermediate probabilities.

There are many other ways to train this mapping, and many techniques from generative modeling are applicable. For example, one approach is to directly match the encoded latent samples by minimizing a reconstruction loss between $z_x$ and $M^{\gets}(z_y)$. In the case that $\ell_x = \ell_y$ have the same dimension, the pushforward of $q_{\phi_y}(z_y \mid y)$ under $M^{\gets}$ can be learned to approximate $q_{\phi_x}(z_x \mid x)$. { Alternatively, KL divergences are also applicable in place of the MSE loss we use.}

The vsPAIR framework, in principle, offers many advantages--in addition to sparse representations and UQ-aware estimation, the vsPAIR framework does not require access to the forward operator. Regarding computational cost, vsPAIR remains competitive against other methods for inference, requiring only a forward pass through the observation encoder, the latent mapping $M^\gets$, and finally the QoI decoder to obtain a reconstruction.  The tradeoff, however, is an expensive offline learning phase as training can be less stable and require more epochs to converge due to the sparsity constraints and the variational structure. In contrast, iterative UQ methods such as diffusion posterior sampling \cite{chung2023diffusion, song2021solving, kawar2022denoising} have simpler training but incur repeated network evaluations at inference, as each sample requires a full iterative trajectory.

At the core of the vsPAIR architecture is the interplay between sparsity and UQ. We posit that the combination of probabilistic structure and sparse encoding offers benefits beyond their individual contributions.  

First, the paired architecture enables \emph{interpretable uncertainty}. By learning separate latent representations for the QoI and observation, the QoI encoder learns features anchored to clean data rather than directly from corrupted observations. This separation of concerns means that latent dimensions encode properties of the underlying signal itself, not artifacts of the measurement process. When uncertainty arises in the latent mapping, it can be attributed to meaningful factors of the QoI. This interpretability is valuable for downstream decision-making, particularly in applications where users must understand what the model is uncertain about, not merely how much.

Second, sparse encoding yields \emph{structured uncertainty}. In a dense VAE, uncertainty diffuses across all latent dimensions in an entangled manner. When sampling from the approximate posterior and decoding, one observes variation in reconstructions but cannot easily attribute this variation to specific factors. With sparse codes, information concentrates into a small number of active dimensions, each corresponding to a dominant factor of variation in the QoI. This concentration allows uncertainty to be traced to specific indices rather than dispersed across the full latent space. 

Third, the spike-and-slab posterior naturally decomposes uncertainty into two components: \emph{which} dimensions are active (captured by the gating probabilities $\omega_i$) and \emph{what values} the active dimensions take (captured by the slab parameters $\mu_i, \sigma_i$). By construction, this decomposition distinguishes structural uncertainty (whether a factor is present) from parametric uncertainty (the factor's magnitude given presence), a distinction unavailable in standard VAEs where small latent values are ambiguous between true inactivity and mere proximity to zero.

{ Note, the vsPAIR framework does not leverage the forward process and the latent posteriors are shaped primarily by how the VAEs learn their approximate posteriors with respect to the latent priors.  Hence, the UQ properties of vsPAIR therefore reflect how the latent representation of the QoI is organized, rather than a calibrated response to observation noise or $p(x \mid y)$ in the Bayesian sense. Coupled with the sparsity mechanism, this structure isolates a small set of active latent dimensions whose variation drives the spread in the reconstruction, allowing the reported uncertainty to be traced to identifiable latent features. 
}

{  Although theoretical calibration with the true posterior is limited, the structure of the framework (\Cref{fig:Bayesnet}) can be leveraged to characterize the mean and covariance of the model-induced $q(z_x \mid y)$ distribution in a reduced setting.  This offers some alignment with $p(x \mid y)$, as it characterizes the distribution of latent codes for $x$ conditioned on $y$. For simplicity, we assume that the QoI encoder, $q_{\phi_x}(z_x \mid x)$, is Gaussian (i.e., a standard VAE) and make the following regularity assumptions:

{
\begin{enumerate}
    \item $\bbE[X \mid Y = y]$ and ${\rm Cov}(X \mid Y = y)$ are both continuous as functions of $y$.
    \item The variational posterior $q_{\phi_x}(z_x \mid x)$ is trained so that  $Z_x$ is conditionally independent of $Y \mid X$. 
    \item The mapping from $x$ to the latent mean is affine, $\mu_x(x) = Bx + c$, and the latent covariance mapping is constant and symmetric positive definite, $\Sigma_x(x) = \Sigma_{Z_x}$.
\end{enumerate}
}
The first condition is an assumption on the true data distribution, while the second inherent to the paired architecture. The third depends on the networks selected for the variational posterior of the QoI encoder and is the most restrictive assumption. \Cref{lemma:main} establishes an expression for the model-induced conditional mean and covariance of $Z_x \mid Y$.

{
\begin{theorem}\label{lemma:main}
    Suppose that Assumptions 1, 2, and 3 are satisfied.  Then, the conditional mean and covariance of $Z_x \mid Y$ induced by the paired encoding framework exist as continuous functions of $y$. In particular, 
    \begin{align}
    \bbE[Z_x \mid Y = y]= B\bbE[X \mid Y = y] + c
    \end{align}
    and
    \begin{align}
        {\rm Cov}(Z_x \mid Y = y) = \Sigma_{Z_x} + B {\rm Cov}(X \mid Y = y)B^\top
    \end{align}
which is symmetric positive definite for all $y$.
\end{theorem}

The expressions follow directly from the tower property of conditional expectations and the law of total covariance \cite{blitzstein2019} (see \Cref{sec:thm1pf}) and can be generalized to more complex network architectures.  The utility of this characterization is that it shows that the mapping from $y$ to desired QoI latent posterior parameters is continuous as a function of $y$, hence it can be represented under the Universal Approximation Theorem assuming sufficient network expressiveness and sufficient regularity on the expressions above \cite{HORNIK1989359}.  This can be used to inform the training of the latent map to obtain more calibrated UQ properties, however, the shape of $q(z_x \mid y)$ remains unknown. Further theoretical analysis and augmentation of the framework to better suit calibrated UQ remains a subject of future research. The extent to which the hypothesized properties are observed in the vsPAIR framework is explored in \Cref{sec:numeric}.
}

}

\section{Numerical Results}\label{sec:numeric}
{We evaluate vsPAIR on a Gaussian baseline with a known posterior, followed by two linear inverse problems: blind inpainting on the MNIST dataset followed by computed tomography (CT) on the LoDoPaB-CT dataset \cite{Leuschner_2021}.  Finally we conclude with a nonlinear inverse problem, where we infer the initial state of a heat equation evolution simulation.  Each experiment highlights different strenghts and limitations of the vsPAIR approach.} \emph{To ensure reproducibility, elementary code will be made available on Github upon initial acceptance}.  Details regarding hyperparameter tuning and network architectures for each experiment can be found in \Cref{sec:hyp} and \Cref{appendix:modeldetails}.  {All experiments use deterministic decoders.}

{
\subsection{Gaussian Linear Model}\label{sec:gaus}
As a simple test case, we begin our numerics by considering a simple Gaussian example with a known posterior.  We consider a Gaussian linear model,
\begin{align*}
    y = Ax + \varepsilon
\end{align*}
with $x \in \mathbb{R}^2, x \sim \mathcal{N}(x \mid 0, I_2), y \in \mathbb{R}^4,A \in \mathbb{R}^{4 \times 2}$ and $\varepsilon \sim \mathcal{N}(\varepsilon \mid 0, 0.1I_4)$.  Here the entires of $A$ are randomly sampled from a standard normal distribution and $(x, y)$ are jointly Gaussian with the following characterizations of the joint and conditional distributions
\begin{align*}
p(x, y) &= \mathcal{N}\!\left(
    \begin{bmatrix} x \\ y \end{bmatrix}
    \;\middle|\;
    0,\;
    \begin{bmatrix}
        I_2 & A^\top \\[4pt]
        A   & AA^\top + 0.1\,I_4
    \end{bmatrix}
\right), \\[6pt]
p(x \mid y) &= \mathcal{N}\!\left(
    x \;\middle|\;
    A^\top\!\left(AA^\top + 0.1\,I_4\right)^{-1} y,\;
    I_2 - A^\top\!\left(AA^\top + 0.1\,I_4\right)^{-1} A
\right).
\end{align*}

For training, we generate $1,\!024$ samples of $y$ from the marginal distribution, and 10 corresponding $x$ samples from the true posterior for a total of $10,\!240$ training points. 200 additional samples of $y$ are generated as test samples.  A latent dimension of $8$ is used for both the $x$ and $y$ encoders.

\begin{figure}[htb]
    \centering
    \includegraphics[width=0.8\linewidth]{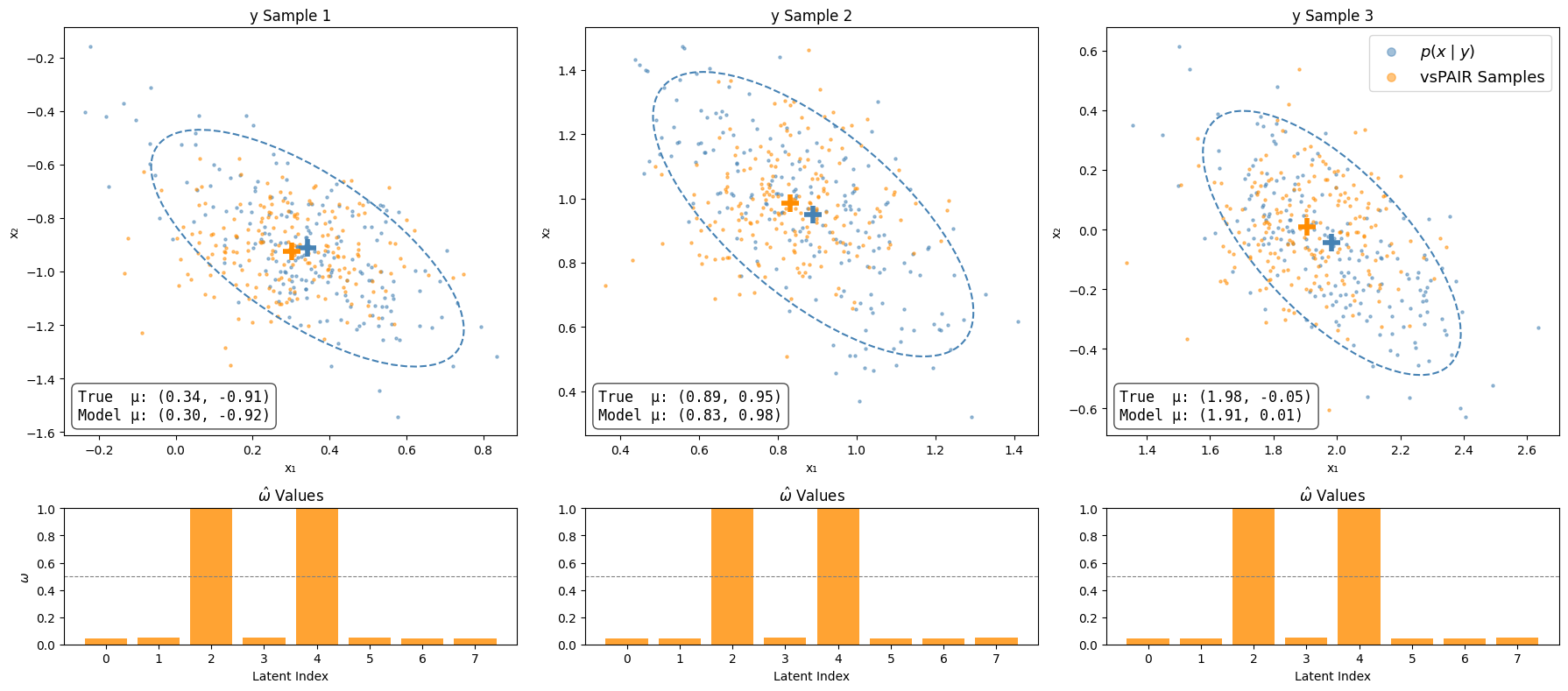}
    \caption{The top image displays 500 samples drawn from the true posterior (blue) with the ellipse representing two standard deviations about the posterior mean (blue cross) for three distinct $y$ in the test set.  500 samples from the vsPAIR mapped distribution are shown in orange alongside the decoded approximate mean, $\hat\omega_x \odot \hat\mu_x$ (orange cross). The values of of $\hat{\omega}_x$ are shown in the bottom image. }
    \label{fig:gauss1}
\end{figure}

\begin{figure}
    \centering
    \includegraphics[width=0.8\linewidth]{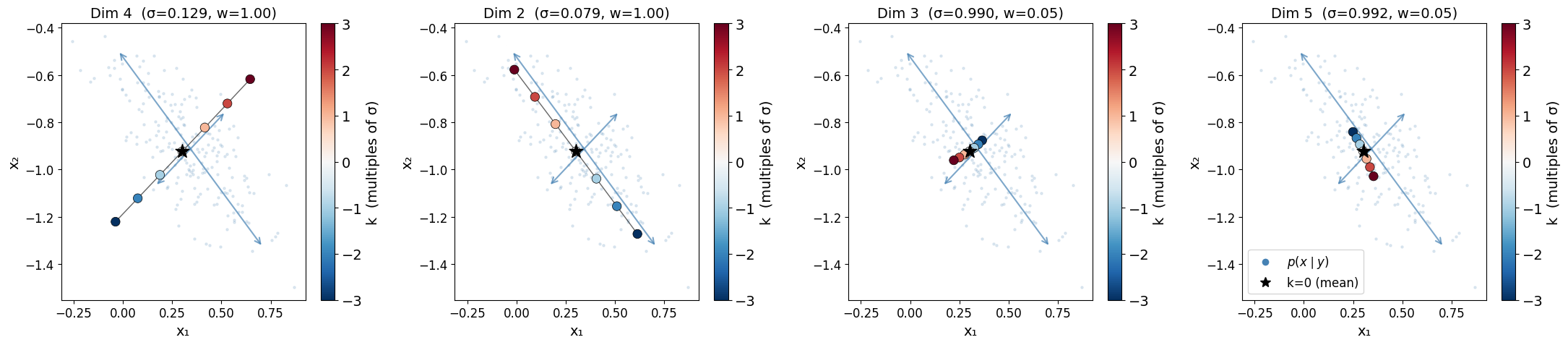}
    \caption{ Samples from the true posterior are displayed alongside the eigenvectors of the true covariance.  Colored points show decoded outputs as a single active latent dimension $k$ is perturbed from $-3\hat\sigma_{x, k}$ to $+3\hat\sigma_{x, k}$ around the predicted mean $\hat\omega_x \odot \hat{\mu}_x$, with all other dimensions  fixed.  This is replicated for the 4 indices with highest values of $\hat\omega_{x, k}$.    }
    \label{fig:gauss2}
\end{figure}

We train vsPAIR on this simple dataset, with details given in \Cref{appendix:GaussDetails}. The results are summarized in \Cref{fig:gauss1}.  Samples from the learned vsPAIR mapped latent distribution generally coincide with samples from the true posterior distribution with nearly identical means.  Moreover, the active latent dimensions collapse into two consistently active dimensions with the remaining dimensions having a near 0 probability of activation across $y$ samples.  Given that the $x$ data are two-dimensional, this behavior is expected and reflective of learning a representative and concentrated subspace to explain the data.  \Cref{fig:gauss2} analyzes the behavior of the reconstruction as the latent indices are varied.  We observe that perturbation of the latent codes along the active dimensions correspond to meaningful variance in the true space: as the active dimensions are perturbed, the reconstructions move away from the mean along directions that are aligned with the eigenvectors of the true posterior covariance.  

Taken together, these results show that vsPAIR is capable of providing a good pointwise estimate for the mean, and concentrating meaningful variance in the reconstruction along a subset of the dimensions.  This indicates that vsPAIR learns a structured latent space, where variance can be attributed to specific features that align with meaningful variance in the reconstruction.  

We emphasize that this setup is simple and limited--standard methods such as PCA or a standard VAE would likely demonstrate comparable properties. The true variance of the posterior is constant in $y$ and vsPAIR is set up for success with calibration of hyperparameters based on prior knowledge of the distribution.  Hyperparamter and network adjustments such as imposing a stronger sparsity constraint can quickly lead to unfavorable results such as collapse into a single dimension. Moreover, a sparsity constraint that is too weak can lead to indices that are consistently active with high variance, yet contribute little to variance in the reconstruction.  Additionally, showing the model multiple $x$ samples for each $y$ pushes the model away from an overconfident pointwise estimate.  These properties are explored in-depth later in more challenging experiments.  Nevertheless, the results of this experiment offer a good baseline study to illustrate that vsPAIR is capable of learning structured latent representations.  The following MNIST experiment offers a more realistic setup that ablates the influence of the paired architecture and sparse latent codes.  }

\subsection{Blind Inpainting on MNIST}\label{sec:MNIST}

Blind inpainting recovers an image from a corrupted observation where both the original content and the corruption mask are unknown. Unlike standard inpainting where missing regions are provided, blind inpainting must simultaneously infer what is missing and what belongs there. This is an ill-posed problem admitting multiple plausible reconstructions.

We construct a dataset from MNIST by deleting 10 randomly placed $5 \times 5$ patches from each $28 \times 28 = 784$ digit image. Patch locations are sampled independently and may overlap, with positions constrained so patches remain within the image boundary. Deleted regions are filled with the minimum pixel value. This yields 54,000 training pairs and 6,000 test pairs of corrupted observations $y$ and clean images $x$. 

To investigate the effects of imposing both sparsity and a probabilistic structure to our inference model, we consider four methods that systematically vary these attributes to isolate their individual contributions and demonstrate their combined effect.

\begin{itemize}
    \item \textbf{PAIR}: Deterministic paired autoencoder with 
    latent dimension 32 for both the observation and QoI.
    \item \textbf{vPAIR}: Variational paired autoencoder with 
    Gaussian encoders, latent dimension 32 for both the observation and QoI, and a learned distributional mapping between latent spaces.
    \item \textbf{sVAE}: Sparse VAE mapping observations directly to QoI with no paired structure and latent dimension 784.
    \item \textbf{vsPAIR}: Sparse VAE with latent dimension 784 for QoI and a standard VAE with latent dimension 32 for observation.
\end{itemize}
A comparison of method properties is provided in \Cref{tab:model_comparison}.

\begin{table}[h]
\centering
\begin{tabular}{l c c c}
\hline
\textbf{Model} & \textbf{Sparsity} & \textbf{Probabilistic} & \textbf{Paired} \\
\hline
PAIR   & \xmark & \xmark & \cmark \\
vPAIR  & \xmark & \cmark & \cmark \\
sVAE   & \cmark & \cmark & \xmark \\
vsPAIR & \cmark & \cmark & \cmark \\
\hline
\end{tabular}
\caption{Comparison of framework properties.}
\label{tab:model_comparison}
\end{table}
All models use the same convolutional neural network architecture and were trained for 100 epochs. Architecture details are provided in~\Cref{appendix:MNISTdetails}.  Moreover, all decoders are deterministic.  

We organize the MNIST evaluation around three questions. First, how does reconstruction quality change as we introduce variational structure and sparsity? Second, do the learned latent representations capture interpretable features, and can individual dimensions be associated with semantically meaningful variation in the reconstructions? Third, does the combination of paired structure and sparsity yield structured uncertainty estimates--that is, can the model not only identify where it is uncertain but also learn latent factors that selectively control the uncertain regions? We address these questions through a combination of quantitative metrics and qualitative analyses across the four methods.

\paragraph{Reconstruction Quality.} \Cref{tab:rec_summary} reports reconstruction error and sparsity metrics across the test set. We tune the sparse models to yield a number of active dimensions comparable to the 32-dimensional dense baselines. This sparsity level is tunable via the Beta distribution regularization parameters $(\alpha_0, \beta_0)$, which is discussed further in \Cref{sec:hyp}.

\begin{table}[h]
\centering
\begin{tabular}{lccccc}
\toprule
Method & $\ell_x \mid \ell_y$ & Avg. nnz & Sparsity & MSE & MSE$_{30}$ \\
\midrule
PAIR & 32 $\mid$ 32 & --- & --- & $\bf{5.76 \times 10^{-2}}$ & --- \\
vPAIR & 32 $\mid$ 32 & --- & --- & $6.61 \times 10^{-2}$ & $\bf{5.39 \times 10^{-2}}$ \\
sVAE & 784 & $\bf{24}$ & $\bf{96.9\%}$ & $8.41 \times 10^{-2}$ & $6.85 \times 10^{-2}$ \\
vsPAIR & 784 $\mid$ 32 & 39 & 95.0\% & $7.62 \times 10^{-2}$ & $6.43 \times 10^{-2}$ \\
\bottomrule
\end{tabular}
\caption{Reconstruction and sparsity metrics on the MNIST blind inpainting test set. $\ell_x \mid \ell_y$: dimensionality of the QoI latent space $\mathcal{Z}_x$ and observation latent space $\mathcal{Z}_y$ for paired methods, and just the latent dimension for sVAE. Avg. nnz: average number of nonzero active dimensions in $\hat{z}_x$ obtained from the observation $y$ for the sparse methods. Sparsity: the percentage of inactive indices. MSE: mean squared error of a single-sample reconstruction $\hat{x}$ against the ground truth $x$. MSE$_{30}$: MSE when the reconstruction is computed as the pixelwise mean of 30 posterior samples.}
\label{tab:rec_summary}
\end{table}

From these metrics, we observe that reconstruction quality decreases as we add variational structure and sparsity constraints: PAIR achieves the lowest MSE, followed by vPAIR, vsPAIR, and sVAE. Note that averaging over 30 posterior samples improves all variational methods, bringing their MSE close to the deterministic baseline. Additionally, the sparse methods achieve approximately 95\% to 97\% sparsity, with 39 and 24 active dimensions for vsPAIR and sVAE on average, respectively. These results are consistent with theoretical expectations: variational structure introduces additional regularization that competes with reconstruction fidelity, consistent with standard VAE findings. We now examine whether these tradeoffs yield meaningful uncertainty estimates.

\paragraph{Qualitative Uncertainty Evaluation.} 

\begin{figure}[htp]
\centering
\includegraphics[width=0.9\textwidth]{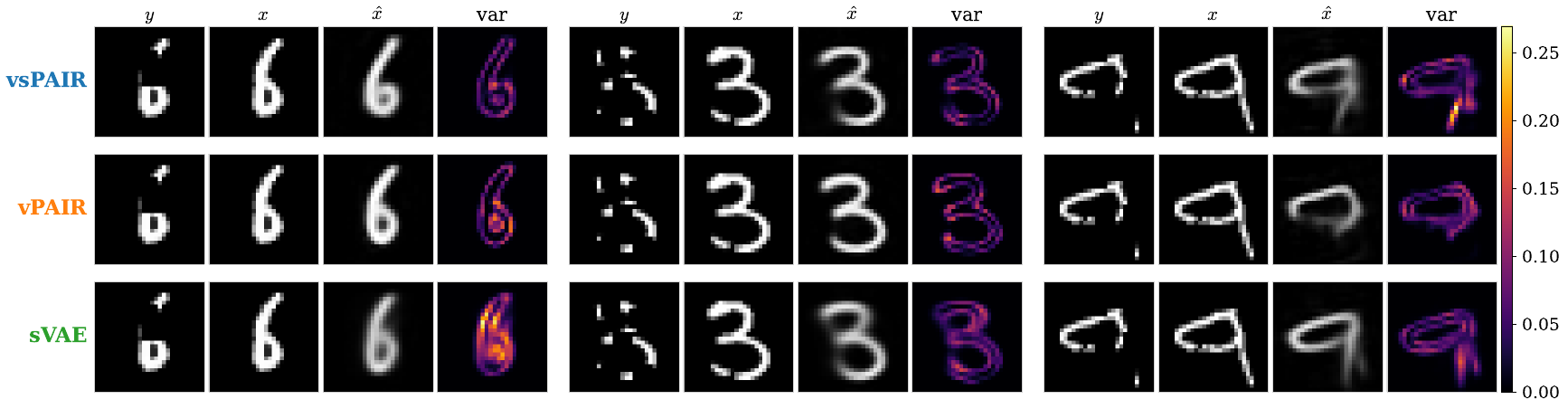}
\caption{Reconstruction and variance comparison across variational methods. Test cases are selected at the 25th, 50th, and 96th percentile of reconstruction error averaged across all three methods, corresponding to easy, medium, and hard cases respectively. For each case, we show the corrupted observation $y$, ground truth $x$, mean reconstruction $\hat{x}$ averaged over 30 samples from the QoI latent distribution, and the pixelwise variance across those samples. vsPAIR is shown on the top row, vPAIR on the middle row, and sVAE on the bottom row. Higher variance in yellow indicates regions where the model produces diverse reconstructions, while lower variance in purple to black indicates confident predictions. All variance maps share a common normalized color scale.}
\label{fig:recon_var}
\end{figure}

We first analyze uncertainty quantification properties of the probabilistic architectures by sampling from the latent distributions. Drawing multiple samples from the predicted latent distribution of the QoI provides insight into where the model is uncertain. In \Cref{fig:recon_var}, we plot the reconstructions for vsPAIR, vPAIR, and sVAE, along with the corresponding variance across multiple sample draws in the latent space. This is computed for three different difficulty levels of images, as measured by the average MSE across the whole image of all three models. 

From this visualization, a difference in how each model handles uncertainty can be observed. The first two images illustrate that in easier cases of reconstruction, all methods are most uncertain around the edges of the digits. The highest difficulty case is most revealing: the corrupted observation could reasonably be a 4, 8, or 9, and the ground truth shows a 9. vsPAIR reconstructs the digit with a noisy mean yet indicates high variance precisely in the tail region. vPAIR reconstructs primarily the known areas of the digit but produces low variance in the unknown regions. sVAE shows the highest variance overall, which is seemingly reflective of its weaker reconstruction quality; its uncertainty estimates appear informative in the hard case but noisier across all difficulty levels.  Whether the sVAE learns structured uncertainty or shows high variance because it cannot reconstruct the images is explored in the proceeding experiments.

\begin{figure}[htp]
\centering
\includegraphics[width=\textwidth]{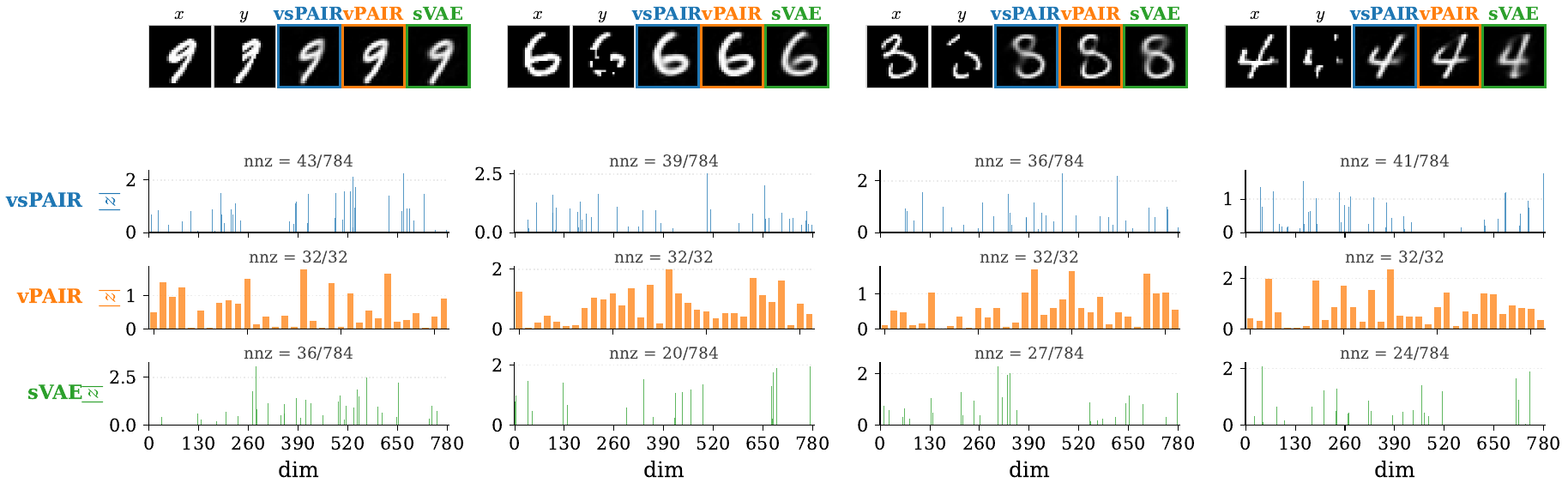}
\caption{Sparse latent representations across methods. Top: corrupted observation $y$, ground truth $x$, and reconstructions from each method. Bottom: absolute value of each latent dimension $|z_j|$ obtained through the inversion path from $y$. Each bar chart is annotated with the number of nonzero dimensions.}
\label{fig:latent_codes}
\end{figure}

To gain further intuition into the learned latent features, we analyze the latent representations directly. \Cref{fig:latent_codes} visualizes the latent representations for four test examples.

\begin{figure}[htp]
\centering
\includegraphics[width=0.9\textwidth]{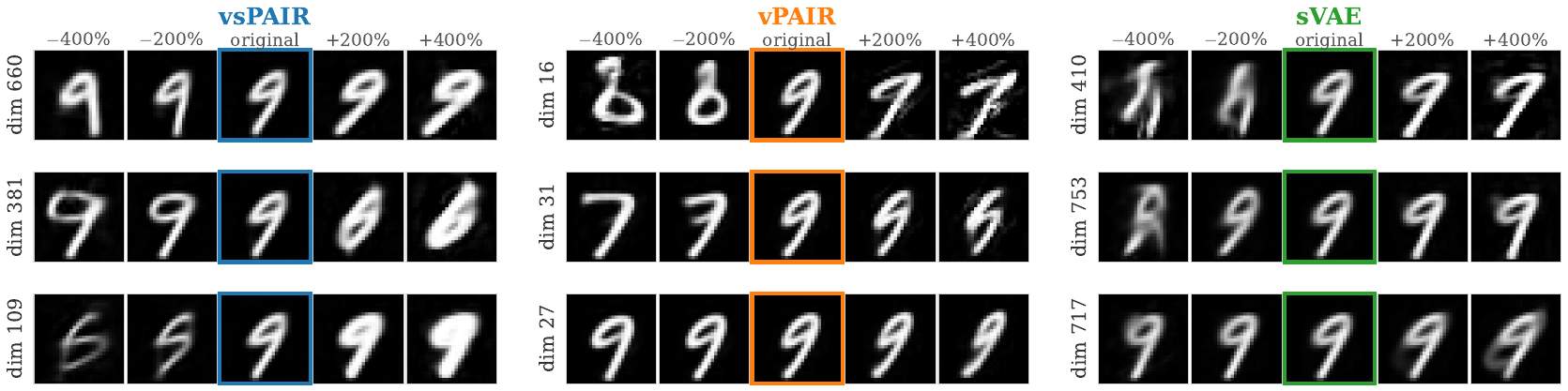}
\caption{Latent dimension perturbation analysis for the first test example in \Cref{fig:latent_codes}. For each method, we select the three latent dimensions with highest sensitivity, measured by reconstruction MSE change under perturbation. For sparse methods, we restrict to dimensions active in all 30 stochastic samples. Each image shows the mean reconstruction across 30 samples with the specified perturbation applied to that dimension. Perturbation ranges from $-400\%$ to $+400\%$ of the original value; the center column in boxes shows the unperturbed reconstruction.}
\label{fig:perturbation}
\end{figure}

The bar charts confirm the sparsity reported in \Cref{tab:rec_summary}: vsPAIR and sVAE activate only a fraction of their 784 dimensions, while vPAIR uses all 32. Using these latent encodings, we examine whether the dimensions encode interpretable features. \Cref{fig:perturbation} takes the first example from \Cref{fig:latent_codes}, the digit 9, and perturbs the three most sensitive latent dimensions for each method, where sensitivity is measured by reconstruction MSE change under perturbation.

The perturbation analysis suggests that all three methods learn dimensions encoding global digit structure. For instance, vsPAIR's dimension 660 appears to control orientation, dimension 381 and 109 interpolate between digit identities, and vPAIR's dimensions 16 and 31 similarly transition through different digits. sVAE's dimensions seem to have less visual interpretability than the other two methods. Notably, vPAIR's third most sensitive dimension seems to show perturbation effects that appear concentrated in the corrupted region rather than globally. This observation raises a natural question: can we systematically identify latent dimensions that selectively control the corrupted regions of the image?

\begin{figure}[htp]
\centering
\includegraphics[width=\textwidth]{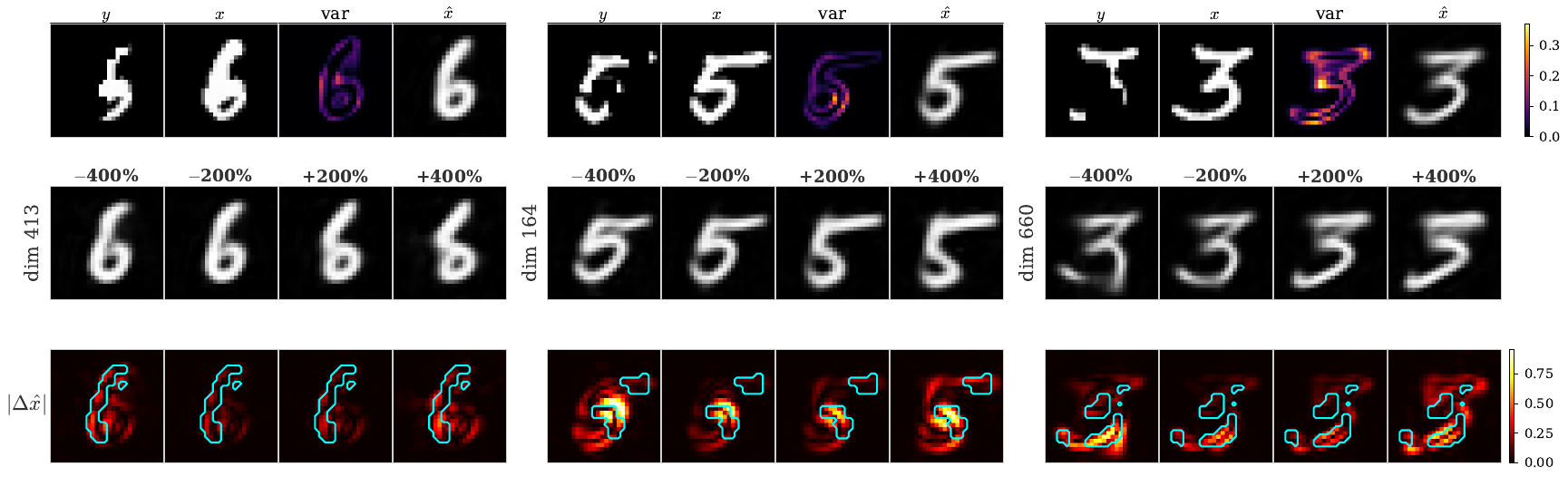}
\caption{Uncertainty-localized latent dimensions in vsPAIR. Each column shows a representative test image selected to illustrate structured uncertainty quantification. Row 1: corrupted observation $y$, ground truth $x$, pixelwise variance, and mean reconstruction $\hat{x}$. Row 2: reconstructions under perturbation of the dimension with highest localization ratio, defined as MSE change in corrupted regions divided by MSE change in uncorrupted regions, ranging from $-400\%$ to $+400\%$. Row 3: absolute reconstruction change $|\Delta\hat{x}|$ with cyan contours indicating the corruption mask.}
\label{fig:uncertainty_localized}
\end{figure}

To investigate this, we search for uncertainty-localized dimensions, defined as latent dimensions whose perturbations disproportionately affect corrupted pixels relative to uncorrupted pixels. To quantify this, we define the \emph{localization ratio:} the ratio of MSE change in corrupted regions to MSE change in uncorrupted regions under perturbation of a single latent dimension. A ratio substantially greater than 1 indicates a dimension whose effects concentrate in corrupted regions rather than affecting the reconstruction uniformly. \Cref{fig:uncertainty_localized} displays results for vsPAIR, showing for each of three representative test images the dimension achieving the highest localization ratio.

For each test image shown, vsPAIR appears to learn at least one latent dimension whose perturbation effects concentrate within the corrupted regions, indicated by cyan contours in Row 3. The digit identity remains stable across perturbations in Row 2: a 6 remains a 6, a 5 remains a 5, while the reconstructed content in the corrupted region varies. This provides evidence that vsPAIR may be able to separate global structure from local uncertainty, with some dimensions controlling what digit is present while others control how the corrupted regions are filled in. Such a decomposition would enable interpretable uncertainty quantification, allowing identification of not just that the model is uncertain, but which latent factors drive that uncertainty and what image regions they affect.

We emphasize that these qualitative findings are inherently difficult to control or predict. There is no principled way to dictate which features an autoencoder learns, and different sparsity levels or hyperparameter choices could yield different learned representations. The specific uncertainty-localized dimensions observed here may not appear in all trained models and all test images. Nevertheless, these observations provide empirical evidence supporting the proposition that vsPAIR can learn structured and interpretable uncertainty representations. 
\paragraph{Quantitative Uncertainty Evaluation.}

\begin{figure}[htp]
\centering
\begin{subfigure}[t]{0.48\textwidth}
    \centering
    \includegraphics[width=\textwidth]{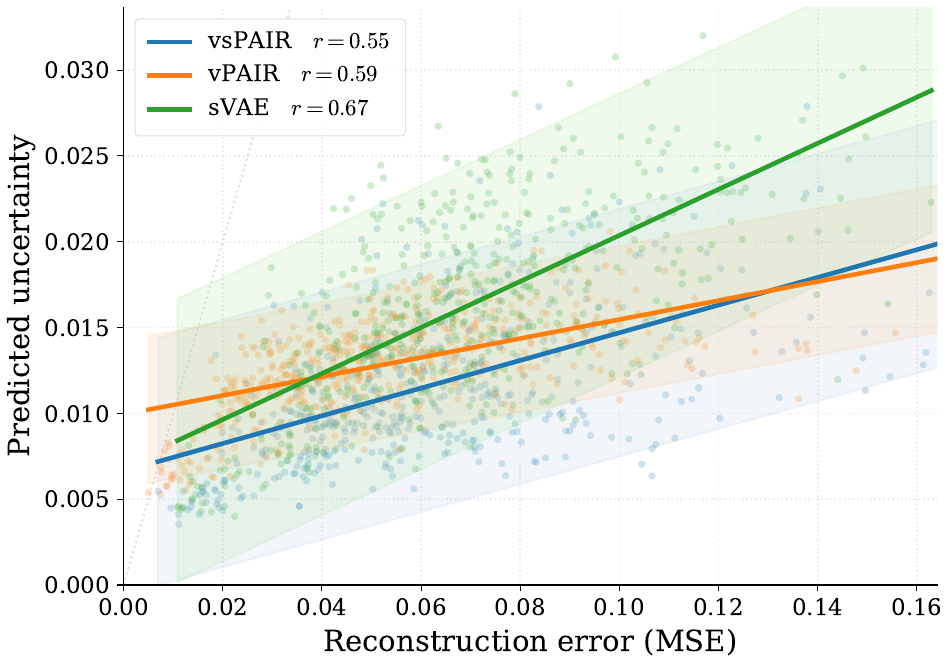}
    \caption{Uncertainty-error correlation. Each point represents one of 500 randomly sampled test images. The x-axis shows reconstruction MSE computed as the mean of 30 samples versus ground truth. The y-axis shows predicted uncertainty as the mean pixelwise variance across those 30 samples. Pearson correlation $r$ is reported for each method; shaded regions indicate 95\% confidence intervals for the linear fit.}
    \label{fig:uq_correlation}
\end{subfigure}
\hfill
\begin{subfigure}[t]{0.48\textwidth}
    \centering
    \includegraphics[width=\textwidth]{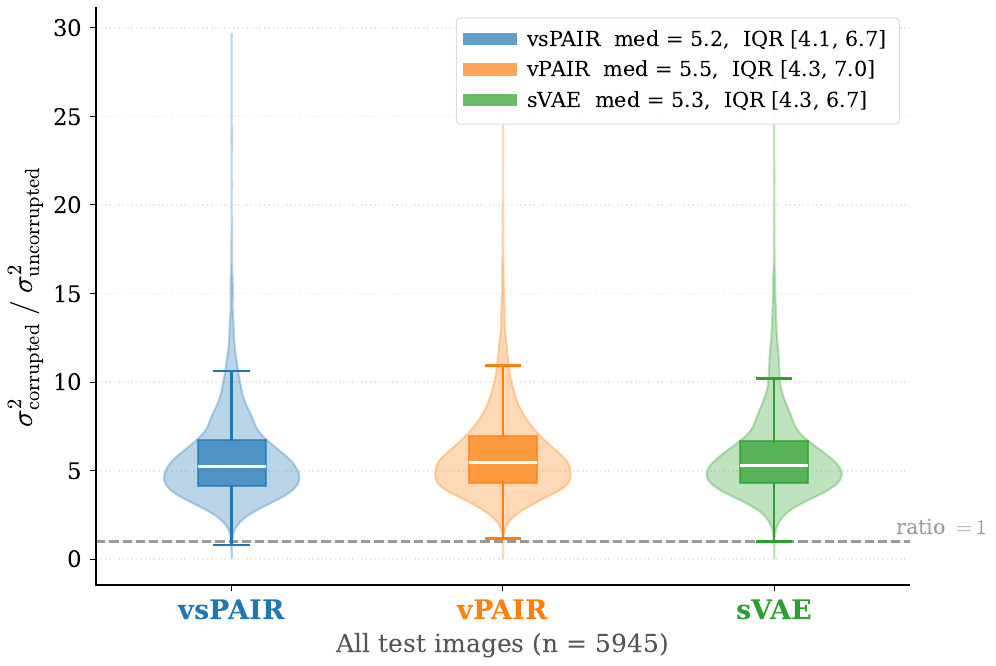}
    \caption{Corrupted-to-uncorrupted variance ratio by region. For each test image, we compute the ratio of mean variance in corrupted regions to mean variance in uncorrupted regions. A ratio above 1 indicates the model places more uncertainty where information is actually missing. Violin plots show the distribution across all test images; boxes indicate interquartile range with median marked in white.}
    \label{fig:uq_localization}
\end{subfigure}
\caption{Quantitative uncertainty evaluation on the MNIST test set.}
\label{fig:quant_uq}
\end{figure}

We provide a quantitative analysis of uncertainty quantification behavior across the variational methods and attempt to confirm the qualitative trends observed in the previous section. \Cref{fig:quant_uq} evaluates two fundamental questions: do the models know when they are wrong, and do they know where information is missing.

\Cref{fig:uq_correlation} asks whether the models know when they are wrong. All three methods show moderate positive correlation between predicted uncertainty and reconstruction error, with $r$ ranging from 0.55 to 0.67. This indicates some awareness of reconstruction difficulty. However, the scale reveals a limitation: predicted uncertainty ranges from 0.005 to 0.03, while reconstruction error ranges from 0 to 0.16. The uncertainty estimates are an order of magnitude smaller than the actual errors, suggesting all models exhibit some degree of overconfidence. This is consistent with the known tendency of VAEs to produce overly narrow posteriors. sVAE shows the highest correlation at $r = 0.67$, though this may reflect its higher overall uncertainty and error rather than better calibration.

\Cref{fig:uq_localization} asks whether uncertainty concentrates in corrupted regions. All methods place approximately $5\times$ more variance on corrupted pixels than uncorrupted pixels, with medians ranging from 5.2 to 5.5 and every test image having a ratio greater than 1. The models correctly identify where information is missing, not merely that information is missing.

Notably, the three methods perform similarly on both metrics. The corrupted-to-uncorrupted variance ratio appears to be a property of variational reconstruction itself rather than a distinguishing benefit of sparsity or paired structure. However, these metrics measure whether uncertainty falls on corrupted pixels, but do not capture whether uncertainty aligns with semantic ambiguity or whether the model learns latent factors that selectively control the corrupted regions.

To address this, we return to the localization ratio introduced in \Cref{fig:uncertainty_localized}, now evaluated systematically across the full test set. For each image and method, we perturb each consistently active latent dimension and identify the dimension achieving the highest localization ratio. We restrict to consistently active dimensions, defined as those active in all 30 stochastic samples, because only stable dimensions permit reliable analysis of learned factors.

\Cref{tab:consistency} reports the number of consistently active dimensions across methods. vPAIR uses all 32 dimensions by construction. vsPAIR maintains an average of 11.5 consistently active dimensions per image, providing a stable subset for analysis. sVAE averages only 2.6 consistently active dimensions due to its stochastic gating, limiting the pool of dimensions available for localization analysis.

\begin{table}[h]
\centering
\begin{tabular}{lccc}
\toprule
 & vsPAIR & vPAIR & sVAE \\
\midrule
Always active (mean) & 11.5 & 32.0 & 2.6 \\
Always active (min) & 0 & 32 & 0 \\
Always active (max) & 18 & 32 & 9 \\
\bottomrule
\end{tabular}
\caption{Consistency of active latent dimensions across the MNIST test set. For each image, we count dimensions active in all 30 stochastic samples. Consistently active dimensions form the basis for the localization analysis in \Cref{fig:localization_histograms}, as only stable dimensions permit reliable attribution of learned factors. vPAIR uses all dimensions by construction, vsPAIR maintains a moderate stable subset, and sVAE shows high variability.}
\label{tab:consistency}
\end{table}

The contrast between vsPAIR and sVAE is notable given that both methods employ sparse encodings. A possible explanation stems from the distributional mapping. vsPAIR predicts gating probabilities that align with the QoI encoder's output, regularizing which dimensions activate for a given observation. sVAE learns its gates directly from corrupted observations without this anchor, potentially resulting in noisier gating decisions. This suggests that sparse encoding and paired structure together may provide both compression and stable factor identification, whereas sparsity alone appears insufficient.

\begin{figure}[htp]
\centering
\begin{subfigure}[t]{0.48\textwidth}
    \centering
    \includegraphics[width=\textwidth]{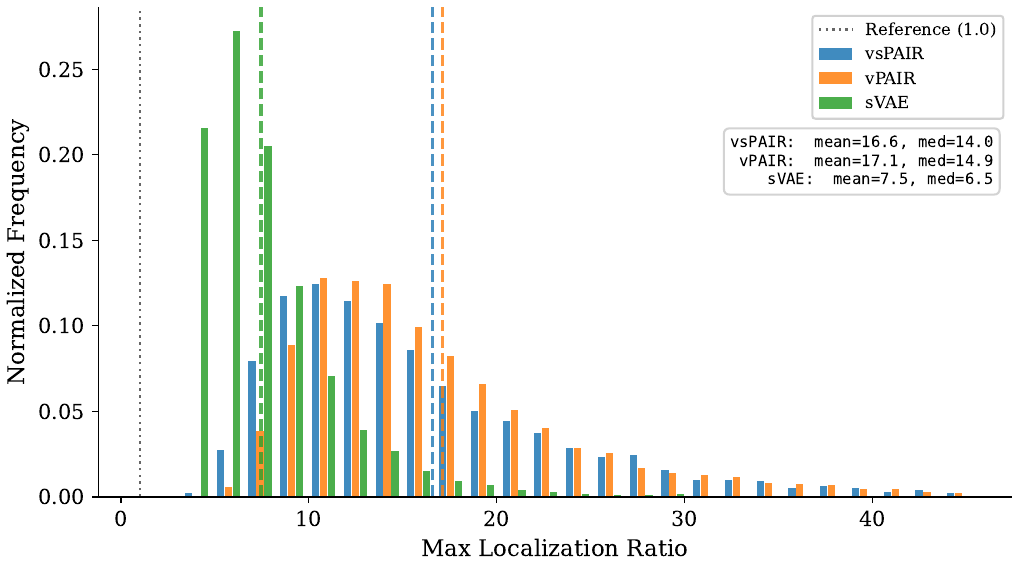}
    \caption{Distribution of maximum localization ratios across the test set. For each image, we identify the consistently active dimension whose perturbation yields the highest ratio of MSE change in corrupted regions to MSE change in uncorrupted regions. Higher values indicate more selective perturbation effects concentrated in corrupted regions.}
    \label{fig:hist_ratio}
\end{subfigure}
\hfill
\begin{subfigure}[t]{0.48\textwidth}
    \centering
    \includegraphics[width=\textwidth]{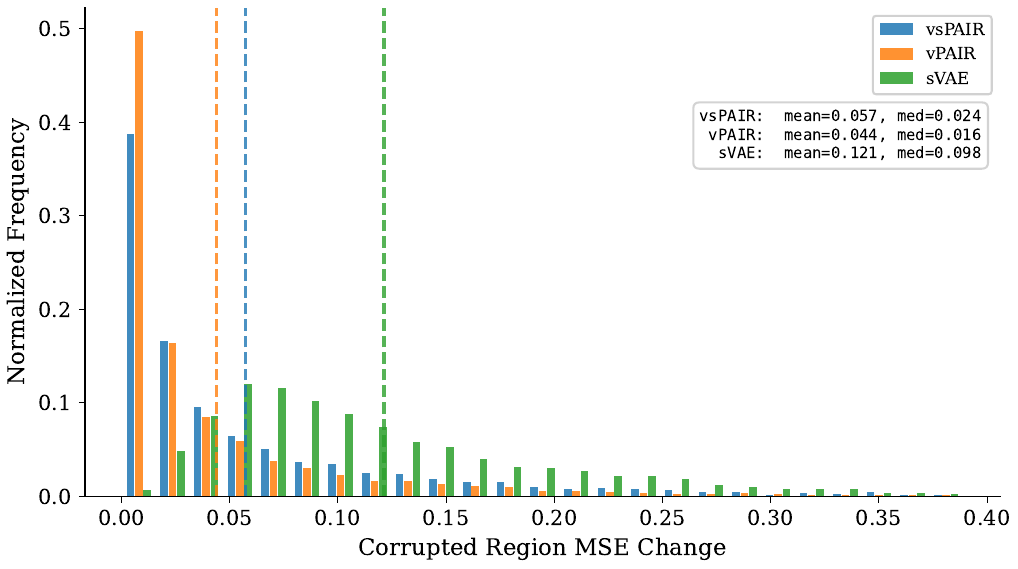}
    \caption{Distribution of MSE change in corrupted regions for the dimension achieving maximum localization ratio. This measures the magnitude of effect that the most localized dimension produces when perturbed, indicating how much control that dimension exerts over the corrupted region content.}
    \label{fig:hist_mse}
\end{subfigure}
\caption{Structured uncertainty quantification across the MNIST test set. Summary statistics are reported in \Cref{tab:localization_stats}.}
\label{fig:localization_histograms}
\end{figure}

\Cref{fig:localization_histograms} shows the distribution of maximum localization ratios alongside the corresponding MSE change in corrupted regions. The histograms suggest separation between the methods, quantified in \Cref{tab:localization_stats}. In terms of selectivity, the paired methods achieve comparable localization ratios with medians near 14 to 15. Perturbations to their most localized dimensions affect corrupted regions roughly 14 to 15 times more than uncorrupted regions. sVAE shows lower selectivity with a median of 6.5, and a substantial fraction of images show ratios below 5. This gap persists even accounting for sVAE's smaller pool of consistently active dimensions: when sVAE does have stable dimensions, they tend to be less selective than those of the paired methods. This provides evidence that the paired structure, which anchors the latent representation to a clean QoI encoder, may be important for learning uncertainty-localized dimensions.

\begin{table}[h]
\centering
\begin{tabular}{llccc}
\toprule
 & & vsPAIR & vPAIR & sVAE \\
\midrule
\multirow{2}{*}{Consistently active only} & Median ratio & 14.0 & 14.9 & 6.5 \\
 & Mean corrupted MSE & 0.057 & 0.044 & 0.121 \\
\midrule
\multirow{2}{*}{All dimensions} & Median ratio & 15.1 & 14.9 & 8.7 \\
 & Mean ratio & 17.7 & 17.1 & 9.7 \\
\bottomrule
\end{tabular}
\caption{Localization ratio statistics across the MNIST test set. The top rows report results when searching only over consistently active dimensions as defined in \Cref{tab:consistency}. The bottom rows report results when searching over all dimensions regardless of gating consistency. The paired methods achieve comparable selectivity in both cases, while sVAE shows lower ratios even when granted access to all dimensions.}
\label{tab:localization_stats}
\end{table}

It is possible that the restriction to consistently active dimensions disadvantages sVAE by excluding potentially selective but unstable dimensions. The bottom rows of \Cref{tab:localization_stats} address this concern. When searching over all dimensions regardless of gating consistency, sVAE improves from median 6.5 to 8.7, but the paired methods also improve slightly to 15.1. The gap does not close, suggesting that sVAE's limitations stem from the representations it learns rather than from the choice of which dimensions to analyze.

\Cref{fig:localization_histograms} further differentiates vsPAIR from vPAIR. Despite comparable selectivity, vsPAIR produces approximately 30\% larger MSE changes in corrupted regions, with mean 0.057 versus 0.044 for vPAIR. A possible interpretation is that this difference reflects how information is distributed across latent dimensions. vPAIR may spread uncertainty-relevant information across all 32 dimensions, diluting the effect of any single perturbation. vsPAIR concentrates this information into its smaller set of consistently active dimensions, approximately 11.5 on average, so each dimension carries more of the total signal. 

sVAE shows the highest corrupted MSE change with mean 0.121, but this likely reflects its lack of reliable representation learning rather than concentrated information. vsPAIR's paired architecture enables separate representation learning: the QoI encoder learns features anchored to clean images, while the distributional mapping learns to predict these features from corrupted observations. This decomposition requires additional parameters, as the paired structure naturally has greater capacity than a single encoder, but this capacity serves to separate concerns rather than simply increase expressiveness. sVAE's single encoder must learn features directly from corrupted observations without this structural prior; sparsity provides compression but not the interpretable decomposition that may arise from the paired design.

Taken together, the qualitative and quantitative analyses provide evidence that vsPAIR can achieve a form of structured uncertainty quantification. The combination of paired structure and sparsity appears to yield latent representations where individual dimensions can both selectively target corrupted regions and produce meaningful effects when perturbed. This comes with a tradeoff: the paired architecture requires additional parameters compared to a single-encoder approach like sVAE. However, this additional capacity serves a specific architectural purpose, enabling separate representation learning for the QoI and observation spaces, which may encourage structured and interpretable latent representations. Sparsity then concentrates information into identifiable factors rather than dispersing it across all available dimensions. Whether this architectural overhead is justified depends on the application; when interpretable uncertainty quantification is a priority, the paired sparse design offers a principled path toward that goal. To illustrate that vsPAIR can be applied to more challenging inverse problems, we consider a CT inversion task.

\subsection{Computed Tomography (CT)}\label{subsec:CT}

{ In this section, we apply vsPAIR to a larger \emph{computed tomography} (CT) example.  For this, we leverage the \emph{Low-Dose Parallel Beam CT} (LoDoPaB-CT) dataset \cite{Leuschner_2021}, consisting of $40, \!000$ scan slides of CT images with simulated measurements.} This dataset uses a subset of existing high-quality CT images from the Lung Image Database Consortium and Image Database Resource Initiative \cite{Armato2011} as ground truth images to simulate corresponding sinograms.  This simulation assumes a parallel beam geometry with 1,000 projection angles, 513 projection beams.

Our experimental setup treats the high-quality CT scan slices as ground truth and seeks to learn a mapping from simulated sinograms to the original scans. Consistent with the vsPAIR framework, we use a sVAE to encode the scan slide (QoI) and a standard VAE to encode the corresponding sinogram (observation). 

For our experiments, we train and test on a subset of the dataset comprising 3,000 training images and $1{,}600$ test images consisting of $800$ unique samples from both the validation and test sets of \cite{Leuschner_2021}.  { The ground truth images are first downsampled by a factor of four and cropped, yielding a $88 \times 88 = 7{,}744$ QoI.  We regenerate observation sinograms using a parallel beam geometry with 248 projection angles and 128 detector pixels.  A linear discrete ray transform is applied to create observations of size $248 \times 128$. Unlike \cite{Leuschner_2021}, we do not simulate Poisson noise and instead add Gaussian noise to each observation, with standard deviation randomly sampled between $0$ and $10\%$ of the sinogram range. }

\begin{figure}
    \centering
    \includegraphics[width=0.4\linewidth]{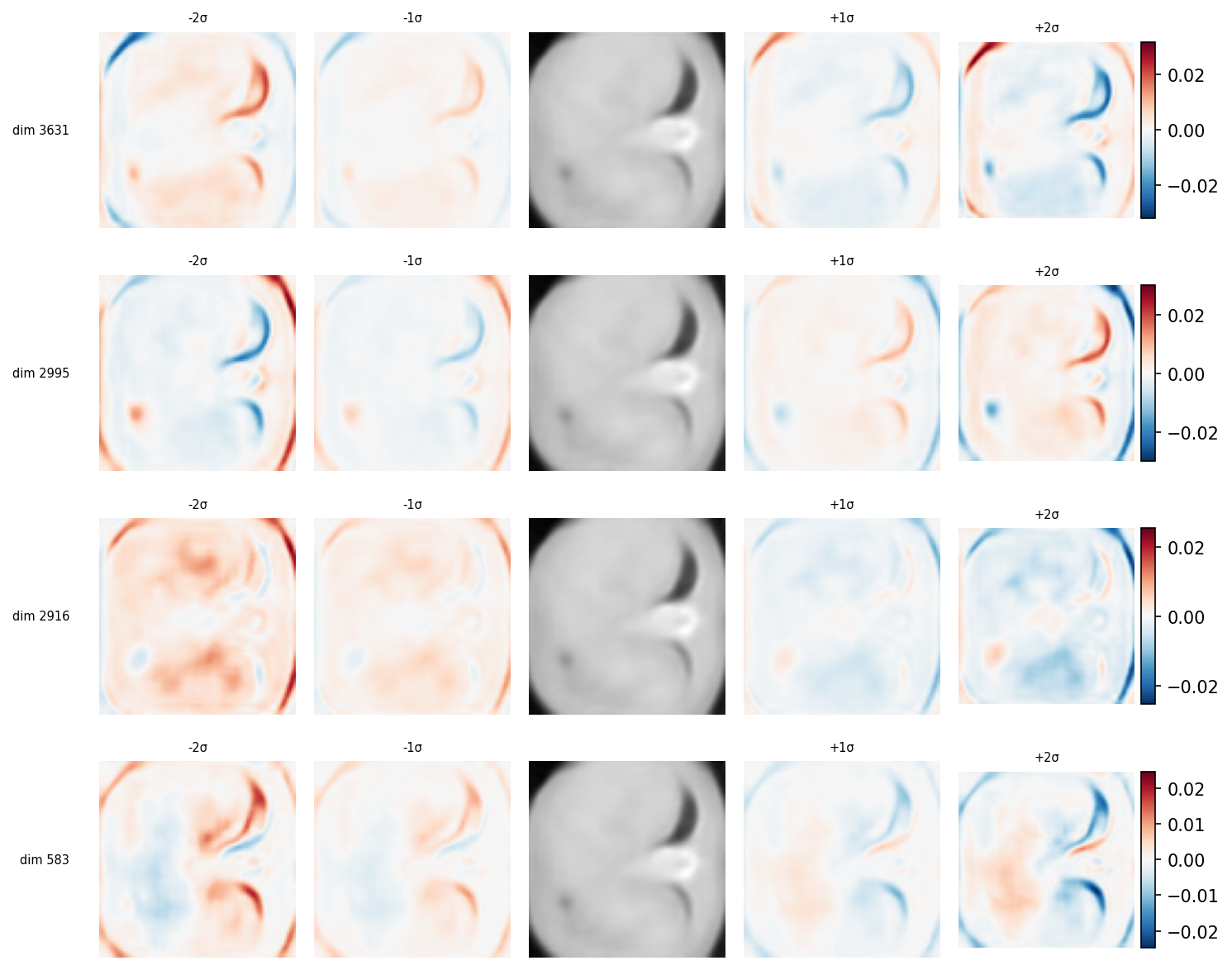} \quad \includegraphics[width = 0.4\linewidth]{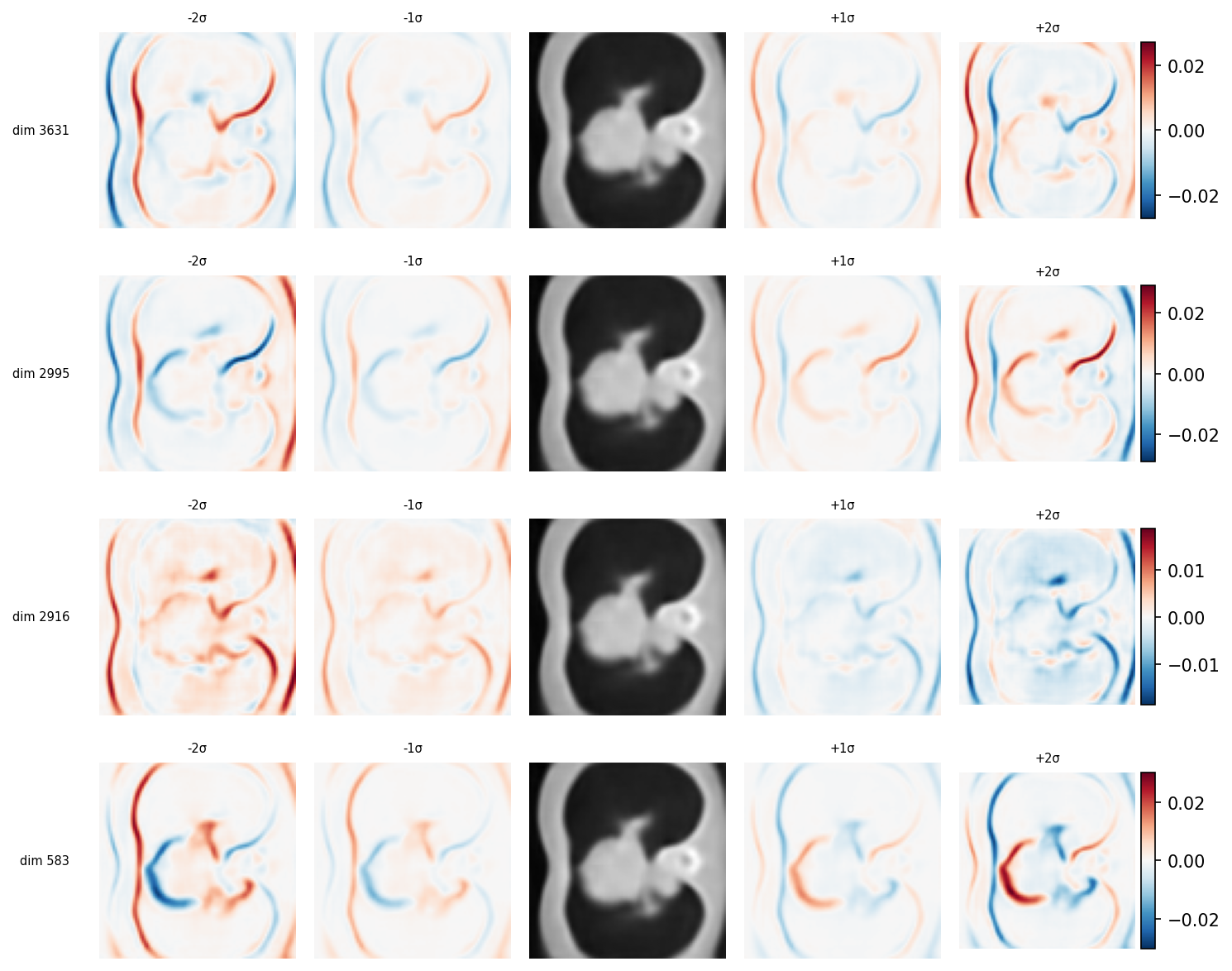}
    \caption{ CT index perturbation study: after fixing the latent representation at the approximate mean, $\hat\omega_x \odot\hat\mu_x$, the four latent indices with average highest variance in $\hat\mu_{x, k}$ across the test set and $\hat\omega_{x,k} > 0.5$ are varied by multiples of $\pm \hat\sigma_{x, k}$ and $\pm2\hat\sigma_{x, k}$ keeping all other indices constant.  These are then decoded, and the pixelwise difference between the resulting image and the mean (center image) is displayed for two images from the test set. }
    \label{fig:ctlattrav}
\end{figure}

{ The vsPAIR model employs CNN-based encoder and decoder architectures, with latent dimensions  $4{,}096$ for the QoI and $1{,}024$ for the observation; here, $\alpha_0 = 1$ and $\beta_0 = 64$.  The results are showcased in \Cref{fig:noise_reconstructions} alongside a total-variation regularized baseline solved using the Alternating Direction Method of Multipliers (TV-ADMM) \cite{10.1561/2200000016}, as well as \emph{Diffusion Posterior Sampling} (DPS) \cite{chung2023diffusion} described in \Cref{subsec:prevwork}.  A detailed description of the DPS implementation, vsPAIR training and hyperparameter details as well as a hyperparameter study for the CT example can be found in \Cref{sec:CTtraining} and \Cref{sec:hyp}.  

\begin{figure}
\centering
\includegraphics[height=0.8\textheight]{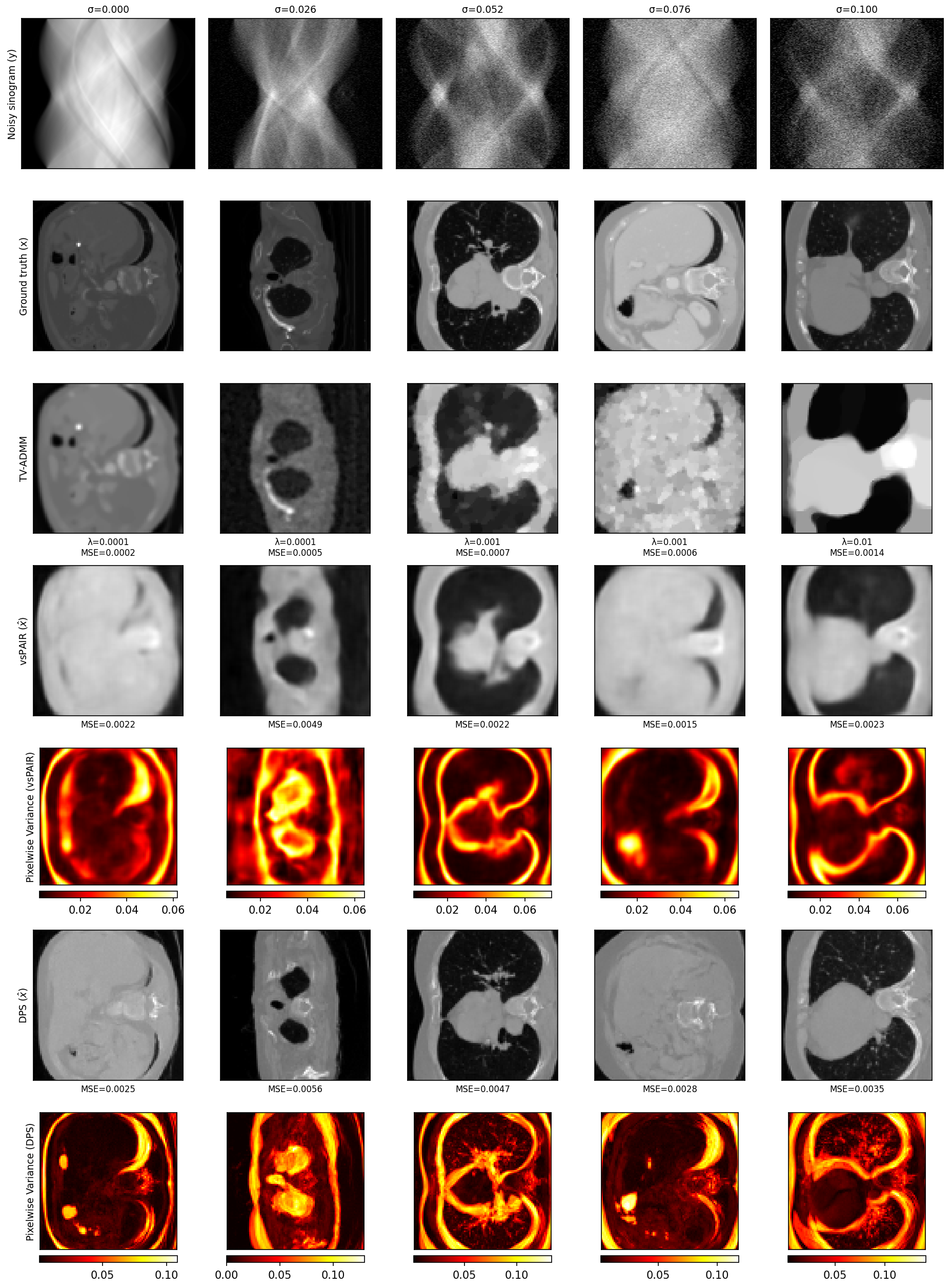}
\caption{ Comparison plot of vsPAIR against TV-ADMM and DPS.  Columns are ordered by increasing noise in the sinogram.  Row (1) shows the noisy sinogram; (2) the ground truth; (3) reconstruction using TV-ADMM with regularization parameter $\lambda \in [10^{-4}, 10^{-4}, 10^{-3}, 10^{-2}, 10^{-1}, 1]$ that gave the best reconstruction measured by MSE; (4) vsPAIR reconstruction (5) vsPAIR pixelwise standard deviation estimated by decoding 50 latent samples; (6) DPS point estimate for a single trajectory (7) DPS pixelwise standard deviation estimated from 10 trajectories.  DPS was implemented with a VP-SDE score model for the QoI prior \cite{song2021scorebasedgenerativemodelingstochastic}--see \Cref{subsubsec:DPSdetails} for further details.  All methods receive the same noise pattern.   }
\label{fig:noise_reconstructions}
\end{figure}

From the \Cref{fig:noise_reconstructions}, we observe that vsPAIR is capable of generating high quality reconstructions with an average MSE of 0.0025 that are competitive with TV-ADMM, while maintaining a sparse representation in the latent space: $1{,}383.3 \pm 32.2$ active dimensions on average.  In particular, vsPAIR is not sensitive to noise, unlike TV-ADMM.  Additionally, drawing multiple samples from the mapped latent distribution reveal features of the image where vsPAIR detects variance.  These are typically concentrated around the edges of the interior cavities, as well as features such as holes or bone structures. This indicates that plausible reconstructions show variation in these details.

While vsPAIR reconstructions capture key attributes of the QoI, they are imprecise, missing fine-grained details.  This coarseness is echoed in the variance maps where uncertainty concentrates on large scale structures such as the edges.  DPS produces sharper reconstructions with uncertainty localized to fine image details.  However, as evident in row 6, DPS reconstructions frequently introduce plausible-looking details that are not present in the ground truth.  Thus, this sharpness can sometimes reflect hallucinated detail rather than recovered signal.  The conservative behavior of the vsPAIR estimates may be favorable in detail critical settings such as medical applications.

The sparse representation of QoI enables the variance in the resulting image to be traced back to distinct latent features. \Cref{fig:density_plot} showcases the distribution of predicted index activation probabilities, $\hat\omega_x$  over all indices and images in the test set.  Mass concentrates around $1$ and $0.3$, indicating that the model selects a subset of indices that are nearly always active, while most have a low-probability of activation.  \Cref{fig:ctlattrav} shows the effect of augmenting the value of the indices where $\hat\mu_x$ is most sensitive to the observation $y$ measured by average standard deviation of $\hat\mu_x$ over the test set.  From this perturbation study, we observe that varying the value of these indices often aligns with variation in dominant modes of the image--for example, in the left image, the second and third index seemingly control the relative size of the hole in the image; in the right image, each index has a strong effect on different edges of the image.  A limitation, however, is that an index having a high probability of activity  does not always imply that it reflects important information and can sometimes be an artifact of how the model trained. Training with a hyperparameter configuration that does not enforce sufficient sparsity can produce indices that are always active, yet contribute little to the reconstruction.

\begin{figure}
    \centering
    \includegraphics[width=0.3\linewidth]{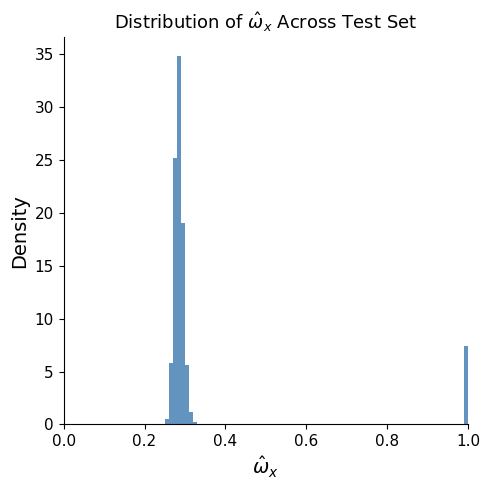}
    \caption{ Histogram of index-wise $\hat\omega_x$ values across the test set for the CT example.}
    \label{fig:density_plot}
\end{figure}

Another limitation of the vsPAIR approach is that the mapped distribution parameters $\hat\sigma_x$ and $\hat\omega_x$ vary little with respect to the observation, $y$: differences in the reconstructions are primarily driven by $\hat\mu_x$. We temporarily emphasize the dependence of the mapped latent distribution parameters on the observation $y$ in our notation for clarity. The standard deviations of $\hat{\omega}_x(y), \hat{\sigma}_x(y),$ and $\hat{\mu}_x(y)$ across $y$ in the test set averaged over the latent dimensions are $0.0080, 0.0092$ and $0.1782$, respectively.  The low variability in $\hat\omega_x(y)$ indicates that the probabilities of each index activating are constant across observations.  This suggests that the model does not learn a unique set of features for each image and has instead identified a reduced subspace to generate reconstructions.  This finding still supports the model concentrating meaningful information along a sparse set of latent values, however, variation in which indices are active as a function of $y$ could offer more observation-specific UQ behavior.  The behavior is likely the effect of our sVAE simplifications.  The original sVAE \cite{tonoliniVariationalSparseCoding2020} utilized a data-driven prior and a spike-and-slab warm-up strategy to enable greater data dependence on the learned activation probabilities. Finally, the vsPAIR model footprint for the CT simulation is large, requiring on the order of 550 million parameters.

}

{ \subsection{Heat Equation}
We conclude the experimental section by exploring application of vsPAIR to a nonlinear inverse problem.  Here, we aim to infer the initial condition of a heat equation simulation, given an observation at a fixed later time.  We consider the following 2D-heat equation with $ u = u(x', y', t)$
\begin{align*}
\partial_t u = \kappa \Delta u, \quad (x', y') \in [0, 2\pi] \times [0, 2\pi],
\end{align*}
periodic boundary conditions and $\kappa = 0.02$. Smooth initial conditions are generated as a finite decaying trigonometric series with a randomly added Gaussian heat source, represented as a discrete $16 \times 16$ grid.   The amplitudes and phases of this series are randomized.  We simulate the evolution from $t = 0$ to $t = 3$, using \emph{Dedalus} with a pseudospectral method and RK222 time integration ($\Delta t = 0.0005$).  This generates QoI, observation pairs $(u( \cdot, \cdot , 0),u( \cdot, \cdot , 3))$. Note that the observation is the simulated value and not the true value of $u$; we write it as $u$ for convenience.

vsPAIR is trained on a train set of 972 QoI observation pairs and validated on 52 held-out samples. An overcomplete latent space is leveraged for both the $x$ and $y$ autoencoders, with latent dimension $512$.  This is paired with a parameter configuration enforcing strong sparsity: $\alpha_0 = 1$, $\beta_0 = 255$.
\Cref{fig:heatrecons} shows the result of inversion on several test samples using vsPAIR.  Generally, these results indicate that the framework is capable of generating high-fidelity reconstructions of the initial state:  the reconstructions capture the grainy sharp variations in the original image that are smoothed out through diffusion  with an average MSE of $0.071$. The pixelwise variation plots over $50$ samples are particularly insightful, as they show that the reconstruction variance concentrates around the primary heat source in each image.   The latent representations remain sparse with $120.6 \pm 3.7$ or $23.7\%$ of the indices active on average.

\begin{figure}
    \centering
    \includegraphics[width=0.8\linewidth]{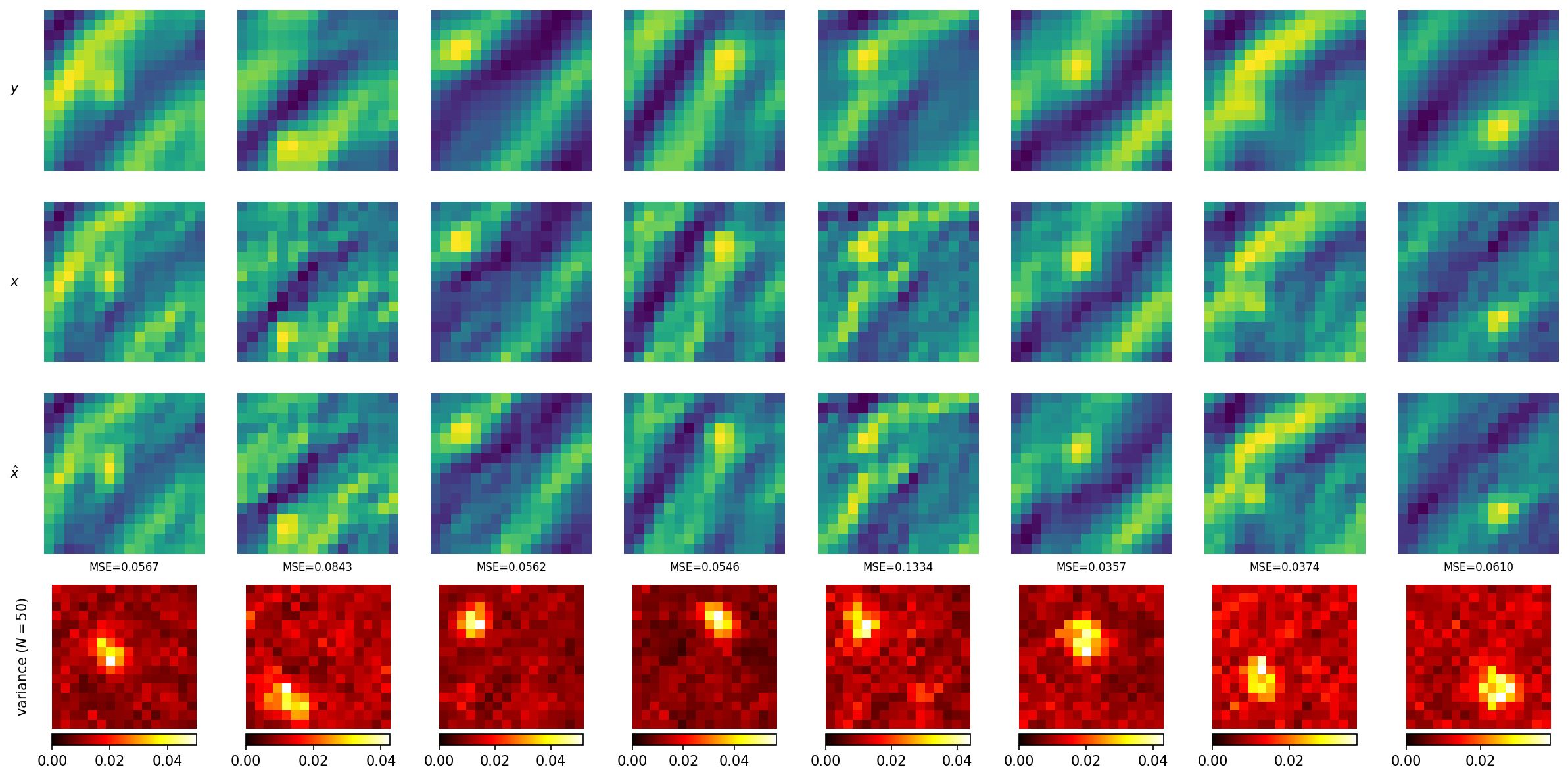}
    \caption{ Heat simulation observations (top row) ($ u(\cdot, \cdot, 3)$) and ground truth (second row) ($u(\cdot, \cdot, 0))$ shown along side vsPAIR reconstruction (third row) and pixelwise variance of 50 samples drawn from the approximate posterior (last row). }
    \label{fig:heatrecons}
\end{figure}

\Cref{fig:uqquant} shows that the model generally assigns higher pixelwise variance to images that are more prone to reconstruction error which suggests that the predicted latent parameters carry information that reflects reconstruction difficulty.  However, similar to the CT example,  $\hat{\omega}_x(y)$ and $\hat\sigma_x(y)$  do not vary significantly with respect to the observation $y$ (standard deviations over the test set averaged across dimensions are $0.0014$ and $0.0078$ respectively). The reconstruction dependency on $y$ is primarily captured in $\hat\mu_x(y)$ with an average standard deviation across the test set of $0.1310$ per dimension.

\begin{figure}[htp]
    \centering
    \includegraphics[width=0.5\textwidth]{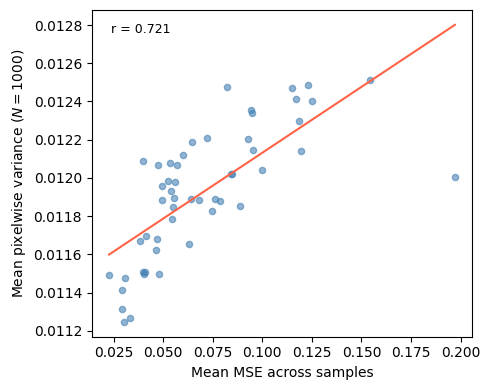}
    \caption{ Mean pixelwise variance of reconstructed samples plotted against mean MSE. Each point represents one test observation, with mean pixelwise variance and MSE estimated from $N = 1000$ samples drawn from the approximate latent distribution. The red line shows a linear fit, with $r = 0.721$ denoting the Pearson correlation coefficient.}
    \label{fig:uqquant}
\end{figure}

\begin{figure}
    \centering
    \includegraphics[width=0.4\linewidth]{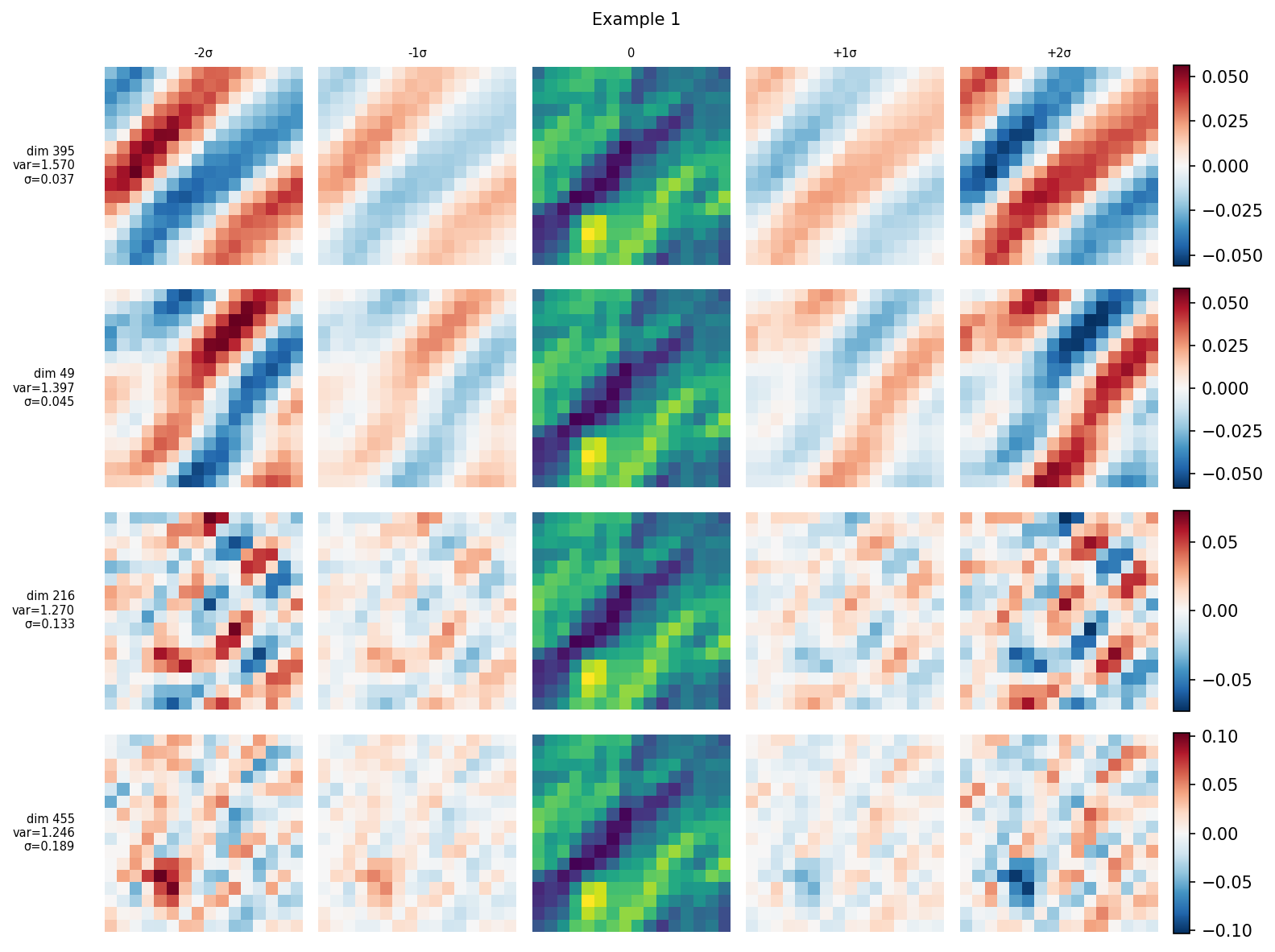}\quad    \includegraphics[width=0.4\linewidth]{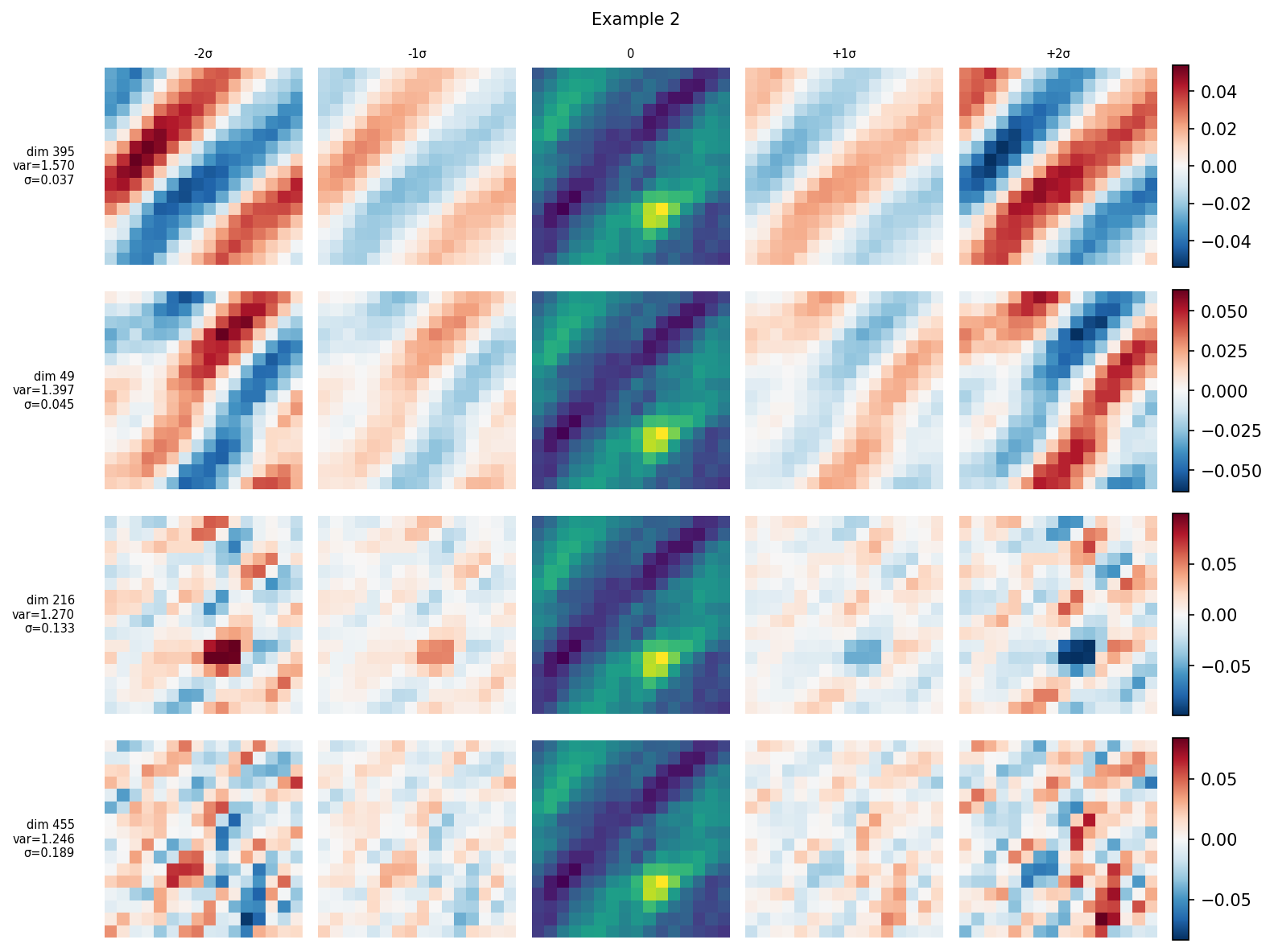}
    \caption{ Heat inversion index perturbation study: After fixing the latent representation at the approximate mean, $\hat\omega_x \odot\hat\mu_x$, the four latent indices with highest variance in $\hat\mu_{x, k}$ across the test set and $\hat\omega_{x, k} > 0.5$ are varied by multiples of $\pm \hat\sigma_{x, k}$ and $\pm2\hat\sigma_{x, k}$ keeping all other indices constant.  These are then decoded, and the pixelwise difference between the resulting image and the approximate mean (center image) is displayed for two images from the test set.}
    \label{fig:heatlattrav}
\end{figure}

We conclude by showcasing latent traversals of the indices with the highest variance in $\hat\mu_{x, k}(y)$.  Perturbation of these indices by their estimated standard deviation $\hat\sigma_{x, k} (y)$ shows a correspondence to macro features of the reconstruction.  Across the two example images, variation of the the first two indices demonstrates similar changes in the reconstruction that seemingly correspond to low frequency characteristics of the QoI.  Although the same dimensions are showcased for both images, the third and fourth dimension correspond to variation in unique image features as it concentrates around the location of the distinct heat sources. This further corroborates vsPAIR's ability to localize information in latent representations.  
}

\section{Discussion and Conclusion}\label{sec:conclusion}

The key contribution of this work is the vsPAIR framework: a data-driven end-to-end framework suitable for fast inference with structured uncertainty. Our framework unifies generative modeling with sparse data representations, yielding low-dimensional latent structures { where reconstruction variance can be traced back to a subset of identifiable features}. The paired architecture encourages interpretability by anchoring QoI representations to clean data, while sparsity provides structure by concentrating information into identifiable factors.  {The underlying mechanism of the encoding process of vsPAIR was formalized.  Future work will expand upon this analysis and further investigate how to calibrate the paired framework with rigorous Bayesian UQ.}  The practical behavior of vsPAIR is assessed empirically.

{ The initial Gaussian experiment provided a limited baseline showcasing that vsPAIR learned variance is structurally meaningful: the active latent dimensions aligned with the dominant directions of variation in the problem, and the sparsity mechanism isolated a small, interpretable set of load-bearing dimensions.} Our results on the MNIST dataset demonstrated that vsPAIR offers suitable reconstructions for ill-posed inverse problems. The introduction of sparsity in the latent space resulted in sparse representations of the QoI, where active indices corresponded to key features of the reconstruction. This, coupled with the probabilistic nature of the VAE and sVAE, can enable structured UQ: an understanding of the model's uncertainty can be generated by drawing multiple samples from the latent distribution, as well as through analysis of the distribution parameters for each likely-active index and how they affect downstream reconstruction. The CT experiments demonstrated that vsPAIR can scale to more complex inverse problems {with competitive reconstructive performance, and variance estimates that align with semantic features of the image.  Finally, the heat initial condition inversion showcased the successful application of vsPAIR to a nonlinear inverse problem.} 

The vsPAIR approach may be considered in applications where interpretable or structured UQ is a priority, particularly those where models can be trained offline but must be deployed efficiently (preferring cheaper inference costs). The framework will generally not match the reconstruction quality of state-of-the-art solvers, so it is best suited to problems where understanding what drives uncertainty is more valuable than maximizing fidelity. Traditional UQ methods provide aggregate measures of predictive confidence, but in many domains the critical question is not merely how much uncertainty exists but what is uncertain. In medical imaging, clinicians require uncertainty estimates that identify which features drive diagnostic ambiguity, enabling informed decisions rather than blind trust in point predictions \cite{huang2024review, abdar2021review}. In geophysical exploration, uncertainty quantification informs drilling decisions and risk assessment; understanding whether uncertainty stems from layer boundaries versus material properties can substantially affect resource management strategies \cite{kotsi2020uncertainty}. For such applications, the ability to trace uncertainty to specific latent factors, and through perturbation analysis to specific regions or features of the reconstruction, may be more valuable than well-calibrated variance estimates alone.  { As an added advantage, vsPAIR does not require access to the forward model of the underlying inverse problem which makes it attractive for settings where the operator is unknown, or too expensive to evaluate.}

These capabilities come at the cost of a large model footprint and the introduction of multiple hyperparameters, though these costs are incurred during training; at inference, vsPAIR requires only forward passes through the network. { As CNNs were generally used for the encoder and decoder, the model footprint is largely dominated by the dense linear heads of CNN networks.  Low rank linear maps can be used to reduce model complexity.}

{ Several further limitations of the vsPAIR method were explored in this work.  The activation probabilities and variance of the predicted latent distributions are largely agnostic to $y$ which reduces vsPAIR's capacity for observation-specific UQ and there is not a clear correspondence between an index being active and its importance in the reconstructions.  As discussed previously, these are likely consequences of our simplifications of the original sVAE.  Nevertheless, the integration of sparsity and paired latent modeling offers a promising foundation for interpretable UQ.  Future work will expand upon the sVAE implementation, as well as alternative approaches to the latent map.  Theoretical developments concern latent-space constraints to enhance interpretability and semantic structure, alongside a more rigorous foundation for UQ estimates.}

\paragraph{Acknowledgments.}  We acknowledge Matthew T. C. Li for early discussions that inspired and clarified the scope of this work.   

\printbibliography

\appendix

{\section{sVAE Implementation Details}\label{appendix:svae}

{
\paragraph{Beta Likelihood Regularization for Adaptive Sparsity.} 

Introducing sparsity into sVAE is not trivial and a sparsity level, $(1 - \rho)$ for the prior must be specified. This can be incorporated index-wise, or fixed for all elements. \cite{tonoliniVariationalSparseCoding2020} originally leveraged a classifier-based approach that builds off of \cite{tomczak2018vaevampprior}, where a classifier is used to pick a specific prior for each input, based on trained pseudo-inputs.  This motivates structure in the latent representations.  

For simplicity, we instead fix the prior probability of activation $\rho$ for all inputs and assume that the distribution of $\rho$ is described by some underlying distribution.  We treat $\rho$ as a learnable parameter regularized by a Beta distribution
\begin{equation}\label{eq:beta_prior}
\rho\sim \mathrm{Beta}(\alpha_0, \beta_0).
\end{equation}
The use of Beta distributions is common in sparse coding \cite{mohamed2012bayesianl1approachessparse, 6618905}.  In particular, our approach draws from \cite{prokhorov2021learningsparsesentenceencoding}, yet is different in that the previous work prescribes a Beta prior to the mixture weights of the spike-and-slab encoder while we enforce it on the sparsity level of the prior.  We consider the negative log-density as a training objective dependent on $\rho$,
\begin{equation}\label{eq:svae_loss_full}
\mathcal{L}_\rho(\rho\, ; \, \alpha_0, \beta_0) = -  \bigl[ (\alpha_0 - 1)\log \rho + (\beta_0 - 1)\log(1-\rho) \bigr].
\end{equation}
Here, $\lambda_\rho$ controls the regularization strength, and terms constant in $\rho$ are dropped. The likelihood term seeks to push $\rho$ towards a probable value under the prescribed loss, but is competing with the sVAE loss term.  This establishes a tradeoff between sVAE performance and sparsity, for which optimization finds a balance.  Since $\rho$ represents the prior probability of a coordinate being active, larger $\rho$ corresponds to denser reconstructions. Setting $\alpha_0 > \beta_0$ yields a left-skewed Beta density that concentrates mass toward higher values of $\rho$, encouraging dense latent representations. Conversely, sparsity is favored when $\alpha_0 < \beta_0$.  Numerical results confirm that this approach aligns with these expectations--see \Cref{sec:hyp}.
}

\paragraph{Bernoulli Relaxation.} Training generative models under sparsity and stochasticity constraints is a well-studied field \cite{louizos2017learning, bengio2013estimatingpropagatinggradientsstochastic}.  A central challenge in training sVAEs arises from the non-differentiable Bernoulli sampling procedure: the loss involves quantities computed from draws of Bernoulli trials. The approach in \cite{tonoliniVariationalSparseCoding2020}, which builds off of \cite{maddison2017concretedistributioncontinuousrelaxation} and \cite{rolfe2017discretevariationalautoencoders}, addresses this issue via a differentiable relaxation of Bernoulli sampling based on a scaled sigmoid and we follow their notation. In particular, the latent encoding is computed element-wise as
\begin{align}
    z_i =T(\eta_i - 1 + \omega_i) \;\cdot\; (\mu_i + \sigma_i \epsilon)
\end{align}
with $\eta_i \sim \mathrm{Uniform}(0,1)$, and $\epsilon \sim \mathcal{N}(\epsilon \mid 0, 1)$.  The expression involving $\epsilon$ is the standard reparameterization trick for VAEs \cite{kingma2013auto}, while the $T(\xi)$ is a differentiable approximation of the step function: ${\bf1}_{\xi \geq 0} (\xi)$, where ${\bf1}(\cdot)$ denotes the indicator function.  This approximation is shifted by $1 - \omega_i$.  If this were a perfect approximation of ${\bf1}_{\xi \geq 0} (\xi)$, for uniform draws $\eta_i$, the outputs of $T(\eta_i - 1 + \omega_i) \sim {\rm Bernoulli} (\omega_i)$.  Critically, the randomness is introduced through $\eta_i$ which is now decoupled from $\omega_i$. 

A Sigmoid function $\sigma(\cdot)$ can be used for $T$ where the input is scaled by a large positive temperature parameter $c$.  As $c \rightarrow \infty$, the scaled sigmoid converges to a step function centered at $0$ with binary outputs. However, for any finite $c$, this relaxation cannot produce exact zeros.

To enforce true sparsity, we leverage a straight-through gradient estimator \cite{bengio2013estimatingpropagatinggradientsstochastic} similar to the approach used in \cite{fallah2022variationalsparsecodinglearned} but without the learned thresholding.  Here, the gradient of the Bernoulli relaxation is used in backpropagation, while thresholding is used to enact exact zeros in the forward pass.  This is implemented as follows:
\begin{enumerate}
    \item Sample uniform noise $\eta_i \sim \mathrm{Uniform}(0,1)$.
    \item Compute a relaxation of the thresholding via the differentiable step function approximation: $s_i = \sigma\bigl(c(\eta_i - 1 + \omega_i)\bigr) \in (0,1)$, where $c > 0$ is a temperature parameter \cite{tonoliniVariationalSparseCoding2020, maddison2017concretedistributioncontinuousrelaxation}. 
    \item Threshold to obtain exact zeros: $\hat{s}_i = \mathbf{1}_{\{s_i > 0.5\}}\in \{0, 1\}$. 
    \item Straight-through gradient estimation: use the mask $\hat s_i$ in the forward pass but use the gradient of $s_i$ for backpropagation
    \cite{bengio2013estimatingpropagatinggradientsstochastic, fallah2022variationalsparsecodinglearned, oord2018neuraldiscreterepresentationlearning}. 
    \item Apply the standard reparameterization ${z}_i^* = \mu_i + \sigma_i \epsilon_i$ with $\epsilon_i \sim \mathcal{N}(\epsilon \mid 0,1)$, then mask: $z_i = \hat s_i \cdot z_i^*$ \cite{kingma2013auto}.
\end{enumerate}

}

\section{Hyperparameter Selection}\label{sec:hyp}

The vsPAIR framework requires tuning several hyperparameters for training.  Parameter estimation for scientific machine learning is an extensive area of research and largely outside the scope of this paper.  However, we provide a limited analysis of how the hyperparameters affect model behavior; more robust statistical analysis and tuning methodologies remain a topic for future studies. {To emulate a realistic setup, we conduct our study on the CT example from \Cref{subsec:CT} with smaller latent spaces: $2{,}048$ and $512$ for the QoI and observation respectively.}

Through extensive trials varying the ratio $\lambda_1 / \lambda_2$, we found { that training remained stable with respect to these parameters.  As a baseline, $\lambda_1$ and $\lambda_2$ were fixed at $1$ and $0.3$ respectively, and $\lambda_b = 0$. } Our first study considers the term $\lambda_\rho \mathcal{L}(\rho ; \alpha_0, \beta_0)$.  The effects of varying the pair ($\alpha_0, \beta_0$), which characterize the Beta distribution, are more interesting than the weight term $\lambda_\rho$ and so we focus our analysis on the former.  Recall that $\rho$ is the probability of an index being active which is pushed towards the maximally-likely value under this parameterization in training. The mean is given by $\frac{\alpha_0}{\alpha_0 + \beta_0}$, and thus the higher the ratio of $\alpha_0$ to $\beta_0$, the closer the mean of this distribution is to $1$.  Thus, we suspect that the magnitude of the ratio $\frac{\alpha_0}{\alpha_0 + \beta_0}$ should influence the observed level of sparsity in the latent representations of $\hat{z}_x$. In particular, a smaller ratio corresponds fewer active indices (sparsity); a larger ratio corresponds to a higher number of active indices (less sparsity).

{We perform a coarse variation of $\alpha_0$ and $\beta_0$ parameters over $[1, 2, 16, 64]$, training each model for 500 epochs with learning rate $0.0001$. We summarize the average MSE in the reconstruction and nnz in $\hat{z}_x$ in \Cref{fig:parammaps}.  Here, we find that the empirical number of active indices matches the trend that we expect: higher $\alpha_0$ to $\beta_0$ ratio regimes induce lower levels of sparsity, and more sparse representations can be obtained by selecting a smaller $\alpha_0$ to $\beta_0$ ratio.  The inversion MSE indicate that the sparsity is not always a good predictor of reconstructive performance.

\begin{figure}[h]
    \centering
    \includegraphics[width=0.8\linewidth]{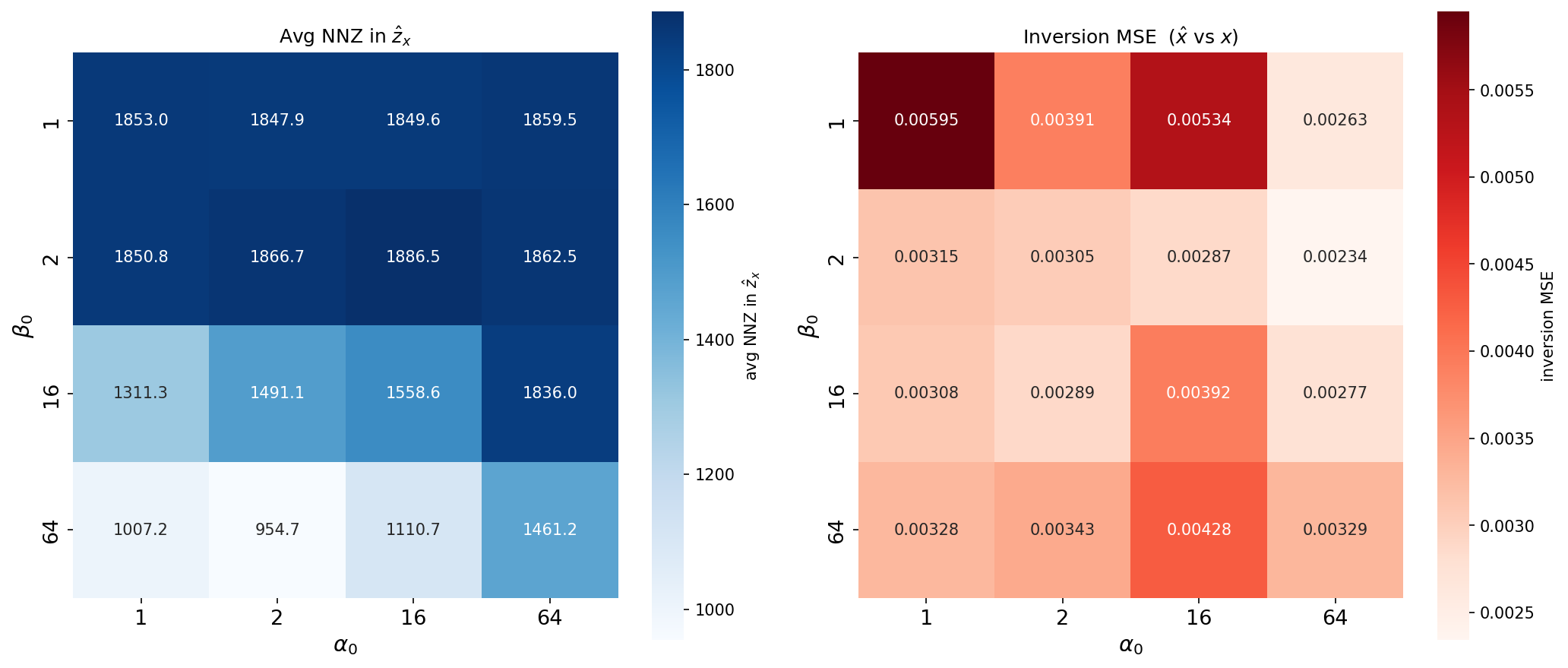}
    \caption{ These heatmaps illustrate the average test set nnz (left) and average number of MSE (right) in the latent representation of $\hat{z}_x$, as $\alpha_0$ and $\beta_0$ are varied over a coarse grid of values $[1, 2, 16, 64]$.}
    \label{fig:parammaps}
\end{figure}

Our next study concerns a hyperparameter not discussed elsewhere in the text. The ELBOs for both the sVAE and VAE can be broken up into a reconstruction and KL term as shown in \Cref{eq:elbo}.  It is standard to assign a hyperparameter weight to the term $\rm{D}_{\rm KL}$ to enforce the stability of training, which is the defining characteristic of a $\beta$-VAE.  The effects of this hyperparameter on VAE training are well studied in other works \cite{burgess2018understandingdisentanglingbetavae, ozcan2025lvaevariationalautoencoderlearnable, pmlr-v97-locatello19a}.  We denote the weight of the KL term in the QoI and observation ELBO as $\gamma_x$ and $\gamma_y$.  Empirically, we found that the weight of $\gamma_x$ has considerable impact on the resulting reconstructions.  Maintaining the same setup as the previous study aide from reducing the number of layers in the encoder/decoder CNNs from 3 to 2, we fix $\alpha_0$ and $\beta_0$ at $1.0$ and $3.0$, respectively, and vary $\gamma_x$ over $[0.001, 0.01, 0.1, 1]$.  The results are shown in \Cref{fig:beta_x_sweep}.

\begin{figure}[t]
\centering
\begin{tikzpicture}

\node (img1) at (0,0)
    {\includegraphics[width=0.7\linewidth]{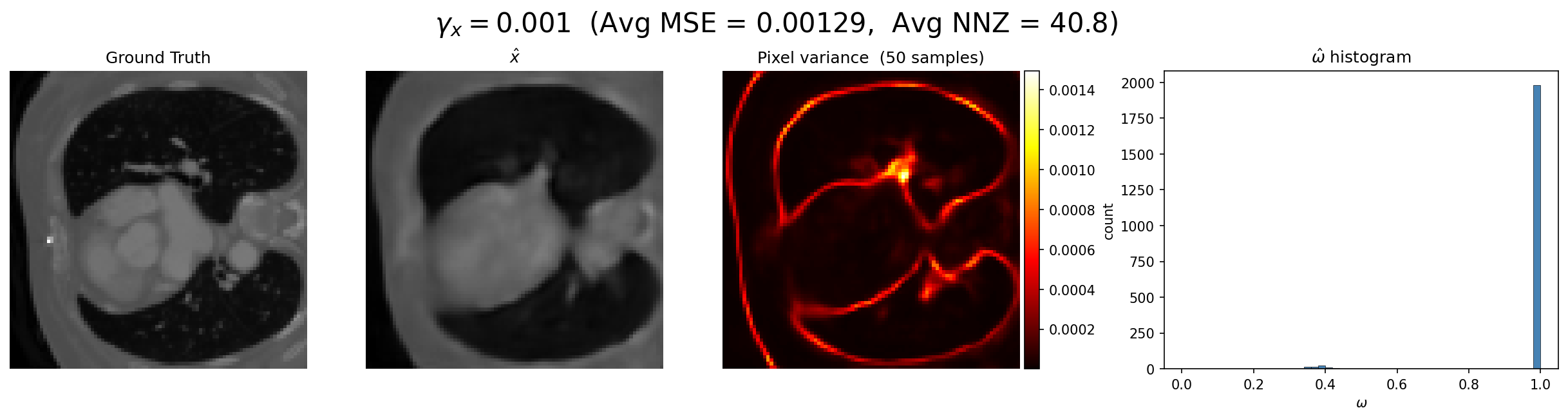}};
\node[anchor=east] at ([xshift=-5pt]img1.west)
    {$\gamma_x=0.001$};

\node (img2) at (0,-3.3)
    {\includegraphics[width=0.7\linewidth]{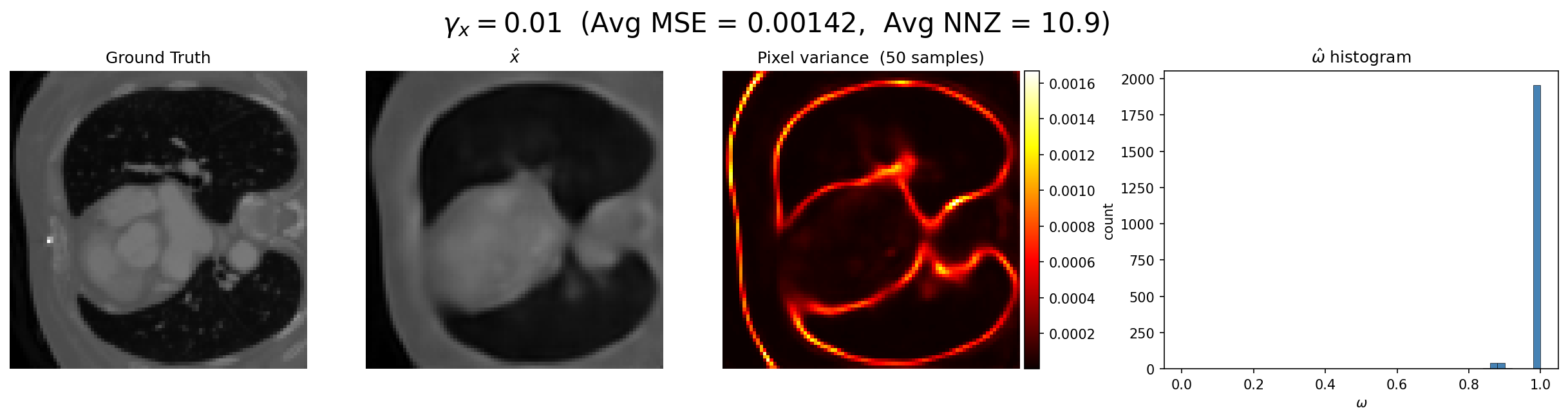}};
\node[anchor=east] at ([xshift=-5pt]img2.west)
    {$\gamma_x=0.01$};

\node (img3) at (0,-6.6)
    {\includegraphics[width=0.7\linewidth]{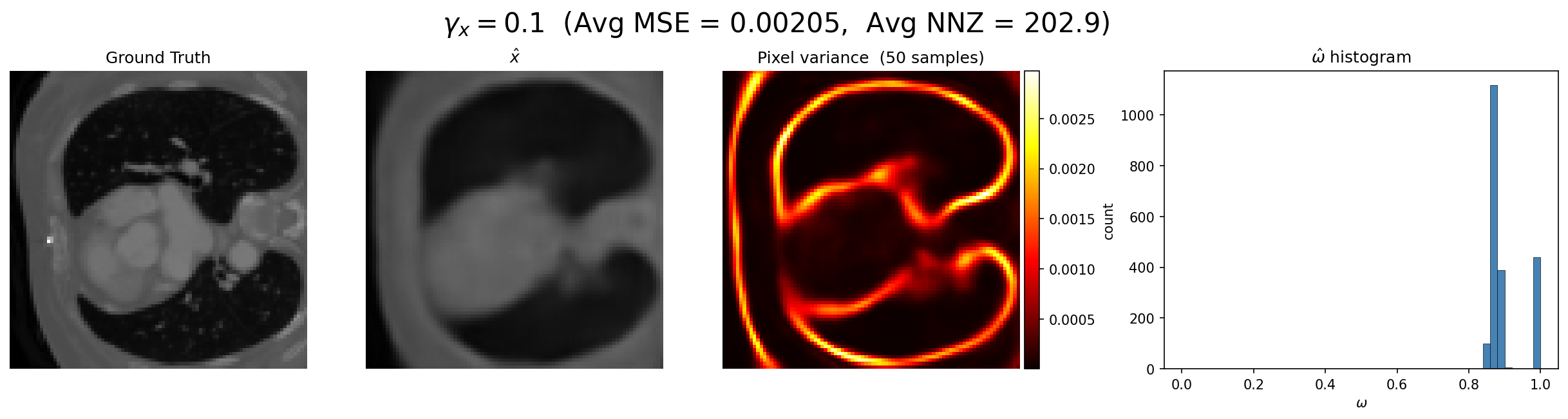}};
\node[anchor=east] at ([xshift=-5pt]img3.west)
    {$\gamma_x=0.1$};

\node (img4) at (0,-9.9)
    {\includegraphics[width=0.7\linewidth]{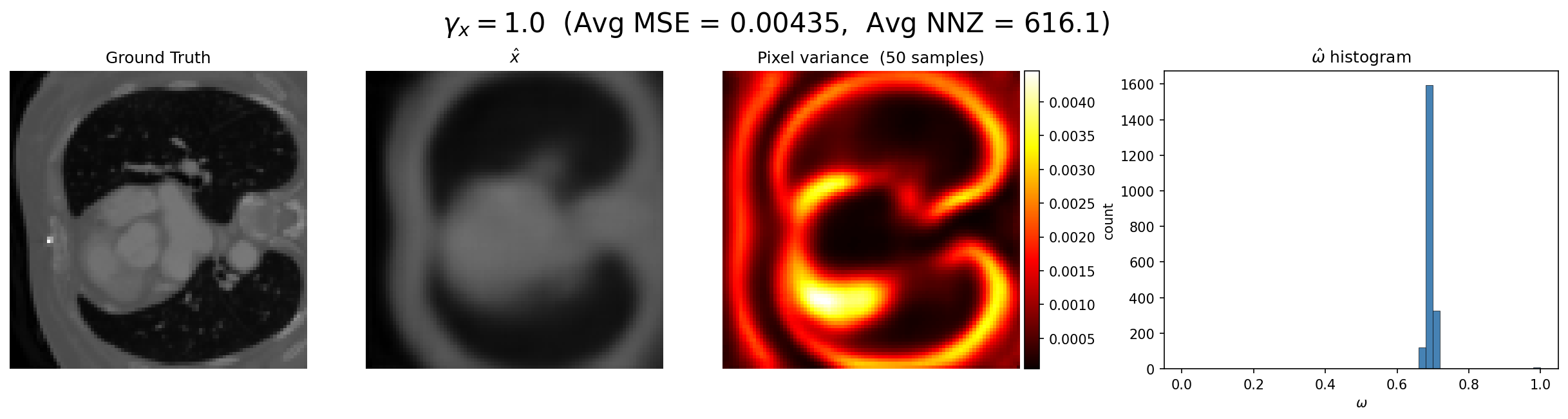}};
\node[anchor=east] at ([xshift=-5pt]img4.west)
    {$\gamma_x=1$};

\end{tikzpicture}

\caption{ Effect of varying $\gamma_x$ over
$[0.001,0.01,0.1,1]$. Images are ordered from top to bottom by increasing $\beta_x$. The left most panel shows an example test-set image followed by the vsPAIR reconstruction and the pixelwise variance across 50 samples of the approximate posterior.  The final figure showcases the histogram of $\hat{w}_x$ for the sample image. }
\label{fig:beta_x_sweep}
\end{figure}

We observe that varying $\gamma_x$ influences both the sparsity structure of the latent space and the resulting reconstruction quality.  For $\gamma_x$ too large, the reconstructions become blurry with a high degree of sparsity in the latent space.  For $\gamma_x$ too small, the reconstructions become sharper, but sparsity is lost in the latent space.  Generally, this parameter must be selected carefully to achieve balance between latent sparsity and reconstruction quality.  As $\alpha_0$ and $\beta_0$ also influence the sparsity level, these three parameters should be tuned jointly.}
\section{Theoretical Results}

\subsection{Proof of Theorem 1} \label{sec:thm1pf}
{

The expression for the conditional mean follows directly from the tower property of conditional expectation and the conditional independence assumption \cite{blitzstein2019}:
\begin{align}
    \bbE[Z_x \mid Y] = \bbE[\bbE[Z_x \mid X, Y] \mid Y]
\end{align}
\begin{align}
    = \bbE[\bbE[Z_x \mid X] \mid Y]
\end{align}
\begin{align}
    = \bbE[\mu_x(X) \mid Y] = B\bbE[X \mid Y] + c.
\end{align}
The last line follows from plugging in the affine expression for $\mu_x(X)$ and the linearity of expectation.

The covariance expression follows directly from the law of total covariance and conditional independence \cite{blitzstein2019}.  From this result, we have
\begin{align}
    {\rm Cov}(Z_x \mid Y) = \bbE[{\rm Cov}(Z_x \mid X, Y) \mid Y] + {\rm Cov}( \bbE[Z_x \mid X, Y  ]\mid Y),
\end{align}
which by the conditional independence gives
\begin{align}
    \bbE({\rm Cov}(Z_x \mid X)\mid Y) + {\rm Cov}(\bbE[Z_x \mid X] \mid Y).
\end{align}
Now, we substitute the encoder mean and covariance, which yields
\begin{align}
    \bbE (\Sigma_x(X) \mid Y) + {\rm Cov}(\mu_x(X) \mid Y),
\end{align}
and simplifies to
\begin{align}
    \Sigma_{Z_x} + B{\rm Cov}(X \mid Y)B^\top,
\end{align}
since $\Sigma_x(x)$ is constant, and $\mu_x(x)$ is affine. This is symmetric positive definite as it is the sum of symmetric positive definite and symmetric positive semi-definite matrices.  Continuity follows from the first assumption.  More general results can be obtained by enforcing more regularity in the parameter maps and the conditional expectation and covariance.}

\section{Model Architectures and Training  Details  }\label{appendix:modeldetails}

{In this section, we provide details on the model architectures and hyperparamters used in each experiment.  All networks were optimized using the Adam optimizer \cite{kingma2017adammethodstochasticoptimization}.}

{
\subsection{Gaussian Example}\label{appendix:GaussDetails}
This section details the network architectures and hyperparameter configure used for the Gaussian example.

\subsubsection{Gaussian Model Details}
The encoder for both $x$ and $y$ are simple MLP networks. The sparse encoder is implemented as 

$$\texttt{Linear}(2 \to 16) \to \texttt{LayerNorm}(16) \to \texttt{SiLU} \to \texttt{Linear}(16 \to 16)$$
then three parallel heads:
$$\texttt{Linear}(16 \to 8) \quad {\mu_x, \log\sigma^2_x, \log\omega_x}.$$
The observation encoder is implemented similarly:
$$\texttt{Linear}(4 \to 16) \to \texttt{LayerNorm}(16) \to \texttt{SiLU} \to \texttt{Linear}(16 \to 16)$$
then two heads:
$$\texttt{Linear}(16 \to 8) \quad {\mu_y, \log\sigma^2_y}.$$
The decoder for $x$ is implemented as
$$\texttt{Linear}(8 \to 16) \to \texttt{LayerNorm}(16) \to \texttt{SiLU} \to \texttt{Linear}(16 \to 16) \to \texttt{LayerNorm}(16) \to \texttt{SiLU} \to \texttt{Linear}(16 \to 2)$$
 and $y$ as
$$\texttt{Linear}(8 \to 16) \to \texttt{LayerNorm}(16) \to \texttt{SiLU} \to \texttt{Linear}(16 \to 16) \to \texttt{LayerNorm}(16) \to \texttt{SiLU} \to \texttt{Linear}(16 \to 4).$$
The latent map has the following architecture:
$$\texttt{Linear}(16 \to 16) \to \texttt{LayerNorm}(16) \to \texttt{ReLU} \to \texttt{Linear}(16 \to 16) \to \texttt{LayerNorm}(16) \to \texttt{ReLU} \to \texttt{Linear}(16 \to 24).$$

As the mean is linear and covariance constant in $y$, nonlinear networks were intentionally used to evaluate if vsPAIR can discover the appropriate latent structure. The entire vsPAIR framework had $3{,}495$ trainable parameters.
}
{\subsubsection{Gaussian Hyperparameters and Training Details}

The hyperparamters used for model training are summarized in \Cref{tab:hyper-gaussian}. Note that the parameters are configured to have mean $1/4$ which corresponds to 2 active dimensions in the latent space.

\begin{table}[h]
    \centering
\begin{tabular}{c | c}
    \hline
    \hline
    Hyperparameter & Gaussian Experiment Value \\
    \hline
    $\lambda_1$ & $1.0$ \\
    $\lambda_2$ & $0.1$ \\
    $\lambda_3$ & $1.0$ \\
    $\lambda_\rho$ & $1.0$ \\
    $\alpha_0$ & $1.0$ \\
    $\beta_0$ & $3.0$ \\
    $\gamma_x$ & $0.05$ \\
    $\gamma_y$ & $1.0$ \\
    $\lambda_b$ & $0$ \\
    \hline
    Training Parameter & \\
    \hline
    nEpochs & 200 \\
    Learning Rate & 1e-3 \\
    Batch Size & 64 \\
    \hline
    \hline
\end{tabular}
    \caption{Enumeration of parameters used in the Gaussian experiment.  $\lambda_1$, $\lambda_2$, and $\lambda_3$ denote the weights for the sVAE ELBO, VAE ELBO, and latent mapping loss, respectively.  $\gamma_x$ and $\gamma_y$, weight the KL divergence terms within each ELBO.  $\lambda_\rho$ weights the Beta distribution loss.  $\alpha_0, \beta_0$ parameterize the Beta density $\mathrm{Beta}(\alpha_0, \beta_0)$ for the learned sparsity level $\rho$.  $\lambda_b$ is the binary entropy penalty weight on predicted gate probabilities.}
    \label{tab:hyper-gaussian}
\end{table}

}

\subsection{MNIST Example}\label{appendix:MNISTdetails}

This section details the network architectures and hyperparameter choices used for the MNIST examples.

\subsubsection{MNIST Model Details}

All models for the MNIST blind inpainting experiment share a common convolutional (CNN) encoder--decoder backbone and differ only in the encoder type, latent dimensionality, and inter-space mapping.  Input images are single-channel $28 \times 28$ with greyscale values normalized between $0$ and $1$.

Each encoder block consists of the following sequence:
\begin{equation}\label{eq:mnist-enc-block}
    \texttt{Conv2D} \to \texttt{BatchNorm2D} \to \texttt{SiLU} \to \texttt{Conv2D} \to \texttt{BatchNorm2D} \to \texttt{AvgPool2D}(2).
\end{equation}
The full encoder network is detailed in Table~\ref{tab:mnist-encoder}.

\begin{table}[h]
\centering
\smallskip
\begin{tabular}{clcr}
\toprule
Layer & Operation & Channels & Output shape \\
\midrule
0 & Input & --- & $1 \times 28 \times 28$ \\
1 & Conv2D ($3 \times 3$) & $1 \to 16$ & $16 \times 28 \times 28$ \\
2 & Equation~\eqref{eq:mnist-enc-block} & $16 \to 32$ & $32 \times 14 \times 14$ \\
3 & Equation~\eqref{eq:mnist-enc-block} & $32 \to 64$ & $64 \times 7 \times 7$ \\
4 & Flatten & --- & 3{,}136 \\
5 & Linear ($\mu$) & $3{,}136 \to \ell$ & $\ell$ \\
6 & Linear ($\log\sigma^2$) & $3{,}136 \to \ell$ & $\ell$ \\
7 & Linear ($\log\omega$) & $3{,}136 \to \ell$ & $\ell$ \\
\bottomrule
\end{tabular}
\caption{CNN network used for the sVAE encoder in the MNIST example.  The last 3 layers correspond to each of the sVAE parameters: $\mu$, $\log\sigma^2$, and $\log\omega$.  For deterministic encoders, only a single linear head is used; for variational encoders, two heads ($\mu$, $\log\sigma^2$) are used.}
\label{tab:mnist-encoder}
\end{table}

The decoders are deterministic and were implemented using a symmetric architecture, where each decoder block replaces the pooling with upsampling:
\begin{equation}\label{eq:mnist-dec-block}
    \texttt{Upsample}(2\times) \to \texttt{Conv2D} \to \texttt{BatchNorm2D} \to \texttt{SiLU} \to \texttt{Conv2D}.
\end{equation}
Further details are outlined in Table~\ref{tab:mnist-decoder}.
\begin{table}[h]
\centering
\smallskip
\begin{tabular}{clcr}
\toprule
Layer & Operation & Channels & Output shape \\
\midrule
0 & Input (Latent Vector) & --- & $\ell$ \\
1 & Linear & $\ell \to 3{,}136$ & 3{,}136 \\
2 & Reshape & --- & $64 \times 7 \times 7$ \\
3 & Equation~\eqref{eq:mnist-dec-block} & $64 \to 32$ & $32 \times 14 \times 14$ \\
4 & Equation~\eqref{eq:mnist-dec-block} & $32 \to 16$ & $16 \times 28 \times 28$ \\
5 & Conv2D ($3 \times 3$) & $16 \to 1$ & $1 \times 28 \times 28$ \\
\bottomrule
\end{tabular}
\caption{CNN network used for the sVAE decoder in the MNIST example.}
\label{tab:mnist-decoder}
\end{table}
All convolutions use kernel size $3 \times 3$, stride 1, and padding 1.

\paragraph{Encoder Variants.}
Three encoder types are used across the model variants:
\begin{itemize}
    \item \textit{Deterministic encoder.}  The flattened features are projected to a latent code $z \in \mathbb{R}^\ell$ by a single linear layer.
    \item \textit{Variational encoder.}  Two linear heads produce $\mu, \log\sigma^2 \in \mathbb{R}^\ell$; samples are drawn via the reparameterization trick $z = \mu + \sigma \odot \epsilon$, $\epsilon \sim \mathcal{N}(\epsilon \mid 0, I)$.
    \item \textit{Sparse variational encoder.}  Three linear heads produce $\mu, \log\sigma^2, \log\omega \in \mathbb{R}^\ell$.  Gate probabilities $\omega_i \in (0, 1]$ are obtained via a clamped exponential $\omega = \exp(-\mathrm{ReLU}(-\log\omega))$.  Samples are drawn using the hard-concrete relaxation with temperature $c = 50$ and straight-through estimation.
\end{itemize}

\paragraph{Model Configurations.}
Table~\ref{tab:mnist-arch} summarizes the architecture of each model variant.  All models were trained for 100 epochs with the Adam optimizer \cite{kingma2017adammethodstochasticoptimization} at learning rate $10^{-4}$ and batch size 64.

\begin{table}[h]
\centering
\smallskip
\begin{tabular}{lccccr}
\toprule
Model & $x$-encoder & $y$-encoder & $\ell_x$ & $\ell_y$ & Parameters \\
\midrule
PAIR   & Deterministic & Deterministic & 32  & 32  & 687{,}042 \\
vPAIR  & Variational   & Variational   & 32  & 32  & 899{,}490 \\
sVAE   & Sparse Var.   & ---           & 784 & --- & 9{,}979{,}106 \\
vsPAIR & Sparse Var.   & Variational   & 784 & 32  & 13{,}067{,}043 \\
\bottomrule
\end{tabular}
\caption{MNIST model architectures and parameter counts.  $\ell_x$ and $\ell_y$ denote the QoI and observation latent dimensions, respectively.}
\label{tab:mnist-arch}
\end{table}

PAIR uses two deterministic encoders and decoders with a linear latent mapping $M^\gets: \mathbb{R}^{32} \to \mathbb{R}^{32}$ (1{,}056 parameters).  vPAIR replaces both encoders with variational encoders and uses a 3-layer MLP with LayerNorm and ReLU activations as the parameter mapper $M^\gets: \mathbb{R}^{64} \to \mathbb{R}^{64}$ (mapping $(\mu_y, \log\sigma_y^2) \mapsto (\hat\mu_x, \widehat{\log\sigma_x^2})$; 12{,}736 parameters; hidden dimension 64).  sVAE is a standalone sparse variational autoencoder that encodes the observation $y$ directly and decodes to $x$, with no paired structure or latent mapping.  vsPAIR uses a sparse variational encoder for $x$ ($\ell_x = 784$), a standard variational encoder for $y$ ($\ell_y = 32$), and a 3-layer MLP parameter mapper $M^\gets: \mathbb{R}^{1{,}568} \to \mathbb{R}^{2{,}352}$ (mapping $(\mu_y, \log\sigma_y^2) \mapsto (\hat\mu_x, \widehat{\log\sigma_x^2}, \hat\omega_x)$; 2{,}644{,}560 parameters; hidden dimension 816).

\subsubsection{MNIST Hyperparameters and Training Details}

\Cref{tab:mnist-hyper} lists the training hyperparameters for each model.
\begin{table}[h]
    \centering
\begin{tabular}{c | c c c c} 
    \hline
    \hline
    Hyperparameter & PAIR & vPAIR & sVAE & vsPAIR  \\
    \hline
    $\lambda_1 $& $1.0 $& $1.0$ & $1.0$ & $1.0$ \\
    $\lambda_2$ & $0.5$& $0.5$ & -- & $0.5$  \\
    $\lambda_3$ & $1.0$ & $1.0$ & -- & $1.0$ \\
    $\lambda_\rho$  & -- & -- & $1.4$ & $1.4$  \\
    $\alpha_0$ & -- & -- & $1.0$ & $1.0$ \\
    $\beta_0$  & -- & -- & $127.0$ & $127.0$ \\
    $\gamma_x$ & -- & $1.0$ & $1.0$ & $1.0$ \\ 
    $\gamma_y$ & -- & $0.1$& -- & $0.1$ \\ 
    $\lambda_b$  & -- & -- & -- & $0.05$\\
    \hline
    Training Parameter & & \\
    \hline
    nEpochs & 100 & 100  & 100 & 100\\
    Learning Rate & 1e-4 & 1e-4 & 1e-4 & 1e-4 \\ 
    \hline
    \hline
\end{tabular}
\caption{Training hyperparameters for the MNIST experiment.   For PAIR, ELBOs are replaced with MSE. }
\label{tab:mnist-hyper}
\end{table}
For the sparse models (sVAE and vsPAIR), the Beta density is set to $\mathrm{Beta}(1, 127)$, which has mean $\frac{1}{128} \approx 0.008$ and concentrates near zero, encouraging high sparsity (i.e., a { low} probability {$\rho$} of each dimension being active).  The gate temperature is $c = 50$ for all sparse models.  PAIR uses only MSE reconstruction losses (no KL terms), with the QoI and observation reconstruction weighted by $\lambda_1$, respectively $\lambda_2$.

{\subsection{CT Example}\label{sec:CTtraining}
This section details the network architectures and hyperparameters used in the CT examples.  The networks generally follow the architectures of the MNIST vsPAIR example.

\subsubsection{CT Model Details}

CNN architectures were used for the encoder and decoder for the CT case study.  Each encoder block consists of the following sequence:
\begin{equation}\label{eq:encoderblock}
    \texttt{Conv2D} \to \texttt{BatchNorm2D} \to \texttt{SiLU} \to \texttt{Conv2D} \to \texttt{BatchNorm2D} \to \texttt{AvgPool2D}(2).
\end{equation}
The full network is detailed in \Cref{tab:var_sparse_encoder}. All convolutions use kernel size $3 \times 3$, stride 1, and padding 1.
\begin{table}[h]
\centering
\begin{tabular}{c c c c}
\hline
Layer & Operation & Channels & Output shape \\
\hline
0 & Input 
  & -- 
  & $1 \times 88 \times 88$ \\

1 & Conv2D ($3 \times 3$) 
  & $1 \rightarrow 16$ 
  & $16 \times 88 \times 88$ \\

2 & \Cref{eq:encoderblock}
  & $16 \rightarrow 32$ 
  & $32 \times 44 \times 44$ \\

3 & \Cref{eq:encoderblock}
  & $32 \rightarrow 64$ 
  & $64 \times 22 \times 22$ \\

4 & \Cref{eq:encoderblock}
  & $64 \rightarrow 128$ 
  & $128 \times 11 \times 11$ \\

5 & Flatten 
  & -- 
  & $15,\!488$ \\

6 & Linear 
  & $15,\!488 \rightarrow 4,\!096$ 
  & $4,\!096$ \\

7 & Linear 
  & $15,\!488 \rightarrow 4,\!096$ 
  & $4,\!096$ \\

8 & Linear 
  & $15,\!488 \rightarrow 4,\!096$ 
  & $4,\!096$ \\

\hline
\end{tabular}
\caption{CNN network used for sVAE encoder in CT example.  The last 3 layers correspond to each of the sVAE parameters: $\mu$, $\log \sigma^2$ and $\log\omega$.}
\label{tab:var_sparse_encoder}
\end{table}

The VAE encoder for the observation data featured an identical structure, yet fitted to the dimensions of the sinogram ($248 \times 148$) and a latent-space of dimension $1,\!024$ with only two linear heads. Both decoders are deterministic and were implemented using a symmetric architecture, where each decoder block replaces the pooling with up-sampling:
\begin{equation}\label{eq:decoderLayer}
    \texttt{Upsample}(2\times) \to \texttt{Conv2D} \to \texttt{BatchNorm2D} \to \texttt{SiLU} \to \texttt{Conv2D}.
\end{equation}
Further details are outlined in \Cref{tab:decoder_x}.
\begin{table}[h]
\centering
\begin{tabular}{c c c c}
\hline
Layer & Operation & Channels & Output shape \\
\hline
0 & Input (Latent Vector) & -- & $4, \!096$ \\

1 & Linear & -- & $15,\!488$ \\

2 & Reshape & -- & $128 \times 11 \times 11$ \\

3 & \Cref{eq:decoderLayer} & $128 \rightarrow 64$ & $64 \times 22 \times 22$ \\

4 & \Cref{eq:decoderLayer} & $64 \rightarrow 32$ & $32 \times 44 \times 44$ \\

5 & \Cref{eq:decoderLayer} & $32 \rightarrow 16$ & $16 \times 88 \times 88$ \\

6 & Conv2D ($3 \times 3$) & $16 \rightarrow 1$ & $1 \times 88 \times 88$ \\
\hline
\end{tabular}
\caption{CNN Network used for the sVAE decoder in the CT example.}
\label{tab:decoder_x}
\end{table}

\Cref{tab:mapnet} outlines the network used for the latent mapping. The entire vsPAIR model had $549,\!708,\!789$ trainable parameters.

\begin{table}[htbp]
  \centering
  \begin{tabular}{clcc}
    \toprule
    Layer & Operation & Shape & Output dim \\
    \midrule
    0 & Input $[\mu_y,\, \log\sigma^2_y]$
      & --
      & $2{,}048$ \\
    1 & Linear
      & $2{,}048 \rightarrow 5{,}120$
      & $5{,}120$ \\
    2 & LayerNorm + ReLU
      & --
      & $5{,}120$ \\
    3 & Linear
      & $5{,}120 \rightarrow 5{,}120$
      & $5{,}120$ \\
    4 & LayerNorm + ReLU
      & --
      & $5{,}120$ \\
    5 & Linear
      & $5{,}120 \rightarrow 12{,}288$
      & $12{,}288$ \\
    6 & Split $[\hat\mu_x,\, \widehat{\log\sigma^2_x},\, \hat w_x]$
      & --
      & $3 \times 4{,}096$ \\
    \bottomrule
  \end{tabular}
  \caption{Architecture for the latent mapper used for the CT experiment.}
  \label{tab:mapnet}
\end{table}

\subsubsection{CT Hyperparameters and Training Details}

All hyper- and training parameters for the CT model showcased in \Cref{subsec:CT} are listed in \Cref{tab:hyper-ct}.

\begin{table}[h]
    \centering
\begin{tabular}{c | c}
    \hline
    \hline
    Hyperparameter & CT Experiment Value \\
    \hline
    $\lambda_1$ & $1.0$ \\
    $\lambda_2$ & $0.3$ \\
    $\lambda_3$ & $1.0$ \\
    $\lambda_\rho$ & $1.0$ \\
    $\alpha_0$ & $1.0$ \\
    $\beta_0$ & $64.0$ \\
    $\gamma_x$ & $0.1$ \\
    $\gamma_y$ & $0.1$ \\
    $\lambda_b$ & $0$ \\
    \hline
    Training Parameter & \\
    \hline
    nEpochs & 1000 \\
    Learning Rate & 1e-4 \\
    Batch Size & 64 \\
    \hline
    \hline
\end{tabular}
    \caption{Enumeration of hyper- and training parameters used in the CT experiment.}
    \label{tab:hyper-ct}
\end{table}

\subsubsection{DPS Implementation}\label{subsubsec:DPSdetails}
An unconditional score model was trained on the clean CT images--VP-SDE \cite{song2021scorebasedgenerativemodelingstochastic} with a linear noise schedule, $\beta(t) = \beta_{\rm min} + t(\beta_{\rm max} - \beta_{\rm min})$ with $\beta_{\rm min} = 0.1$ and $\beta_{\rm max} = 20$.  Images were normalized before training.  The score net was implemented as a U-Net for $88 \times 88$ images with channel widths $(64, 128, 256, 512)$ and three down-sampling $2\times$ stages ($\sim \!32.3$ million trainable parameters).  This was trained for 500 epochs with learning rate $10^{-4}$, using denoising score matching where the network is trained to predict the Gaussian perturbation noise rather than the score directly. The trained score net then provides the prior score for DPS. Posterior sampling was initialized from standard Gaussian noise, and run for $1{,}500$ reverse VP-SDE steps leveraging Euler-Maruyama integration from $t = 1$ to $t = 10^{-3}$.  At each step, the recovered image was un-normalized to physical units for application of the CT forward operator to ensure consistency in the residual computation.  The iterate was then updated using the gradient of the residual, with fixed step size $\zeta = 0.1$.

\subsection{Heat Inversion Example}
This section details the network architectures and parameter choices for the heat inversion experiment

\subsubsection{Heat Inversion Model Details}

The model architecture for the heat experiment is analogous to that of the CT experiment with only 1 encoder and decoder block, adapted to QoI and observations with dimension $16 \times 16$, and a latent space of $512$.  The latent mapper is implemented as 
\[
\begin{aligned}
&\texttt{Linear}(1024 \to 1024) \to \texttt{LayerNorm}(1024) \to \texttt{ReLU} \to \texttt{Linear}(1024 \to 1024) \\
&\to \texttt{LayerNorm}(1024) \to \texttt{ReLU} \to \texttt{Linear}(1024 \to 1536).
\end{aligned}
\]
In total, vsPAIR heat model had $11{,}081{,}065$ trainable parameters.

\subsubsection{Heat Inversion Hyperparameters}
The hyper- and training parameters for the heat experiment are given in \Cref{tab:hyper-heat}.

\begin{table}[h]
    \centering
\begin{tabular}{c | c}
    \hline
    \hline
    Hyperparameter & Heat Experiment Value \\
    \hline
    $\lambda_1$ & $1.0$ \\
    $\lambda_2$ & $0.3$ \\
    $\lambda_3$ & $1.0$ \\
    $\lambda_\rho$ & $1.0$ \\
    $\alpha_0$ & $1.0$ \\
    $\beta_0$ & $255.0$ \\
    $\gamma_x$ & $0.1$ \\
    $\gamma_y$ & $0.1$ \\
    $\lambda_b$ & $0$ \\
    \hline
    Training Parameter & \\
    \hline
    nEpochs & 1250 \\
    Learning Rate & 1e-4 \\
    Batch Size & 32 \\
    \hline
    \hline
\end{tabular}
    \caption{Enumeration of hyper- and trainig parameters used in the heat experiment.}
    \label{tab:hyper-heat}
\end{table}

}

\end{document}